%% file: main.tex
\documentclass{article}
\usepackage{preprint,times}

\input{math_commands.tex}

\usepackage{hyperref}
\usepackage{url}
\usepackage{float}

\usepackage{booktabs}
\usepackage{amsmath,amssymb}
\usepackage{xcolor}
\usepackage{enumitem}
\usepackage{caption,subcaption}
\usepackage{pifont}
\usepackage{wrapfig} 
\newcommand{\cmark}{\textcolor{green!55!black}{\ding{52}}}
\newcommand{\xmark}{\textcolor{red!80!black}{\ding{56}}}

\usepackage[most]{tcolorbox}
\newtcolorbox{promptbox}[1]{%
  breakable, enhanced, colback=black!2, colframe=black!45,
  boxrule=0.5pt, arc=2pt, left=6pt, right=6pt, top=5pt, bottom=5pt,
  fonttitle=\bfseries\small, coltitle=black, colbacktitle=black!8,
  title={#1}, before skip=8pt, after skip=8pt}
\newtcolorbox{promptjson}{%
  breakable, enhanced, colback=white, colframe=black!25,
  boxrule=0.4pt, arc=1pt, left=4pt, right=4pt, top=3pt, bottom=3pt,
  before skip=5pt, after skip=5pt}

\title{SceneBloom: Agentic Indoor Environment\\ Generation with Realistic and Diverse\\ Layouts for Embodied Learning}

\author{Yuhan Liu$^{1}$ \quad Lianhui Qin$^{2,\dagger}$ \quad Shenji Wan$^{1,\dagger}$ \\
$^{1}$New York University \qquad $^{2}$University of California, San Diego \\
\texttt{\{yl10379, sw5973\}@nyu.edu} \qquad \texttt{lianhui@ucsd.edu} \\
$^{\dagger}$Equal advising}

\preprintfinalcopy %

\input{required}

\begin{document}

\maketitle
\vspace{-0.1cm}

\begin{abstract}
Training generalizable embodied agents requires realistic, varied, and simulation-ready indoor environments. Language-guided methods
  support semantic planning and object-level reasoning, but precise spatial grounding remains difficult. Visual generative models offer rich
  priors over plausible arrangements, yet their outputs require geometric reasoning to become physically valid 3D scenes. We introduce SceneBloom, an
  agentic framework that combines visual layout sampling with contract-guided 3D compilation. A hierarchical sampler uses diffusion priors to
  generate furniture arrangements within functional zones. A compiler then uses relations extracted from each proposal to guide geometric fitting
  and local pose correction. Across five room types, SceneBloom attains the highest visual quality and physical stability among the compared methods and is the only method with no detected mesh collisions and out-of-bounds objects. Its layouts are also closer to the distribution of real rooms than those of all compared baselines.
  Object navigation policies trained in these realistic scenes achieve a higher success rate in held-out real scans than those trained on an equal number of real training scans. Code is available at \url{https://github.com/Tinaliu0123/scenebloom}.
\end{abstract}

\vspace{-0.25cm}
\begin{center}
\includegraphics[width=0.98\linewidth]{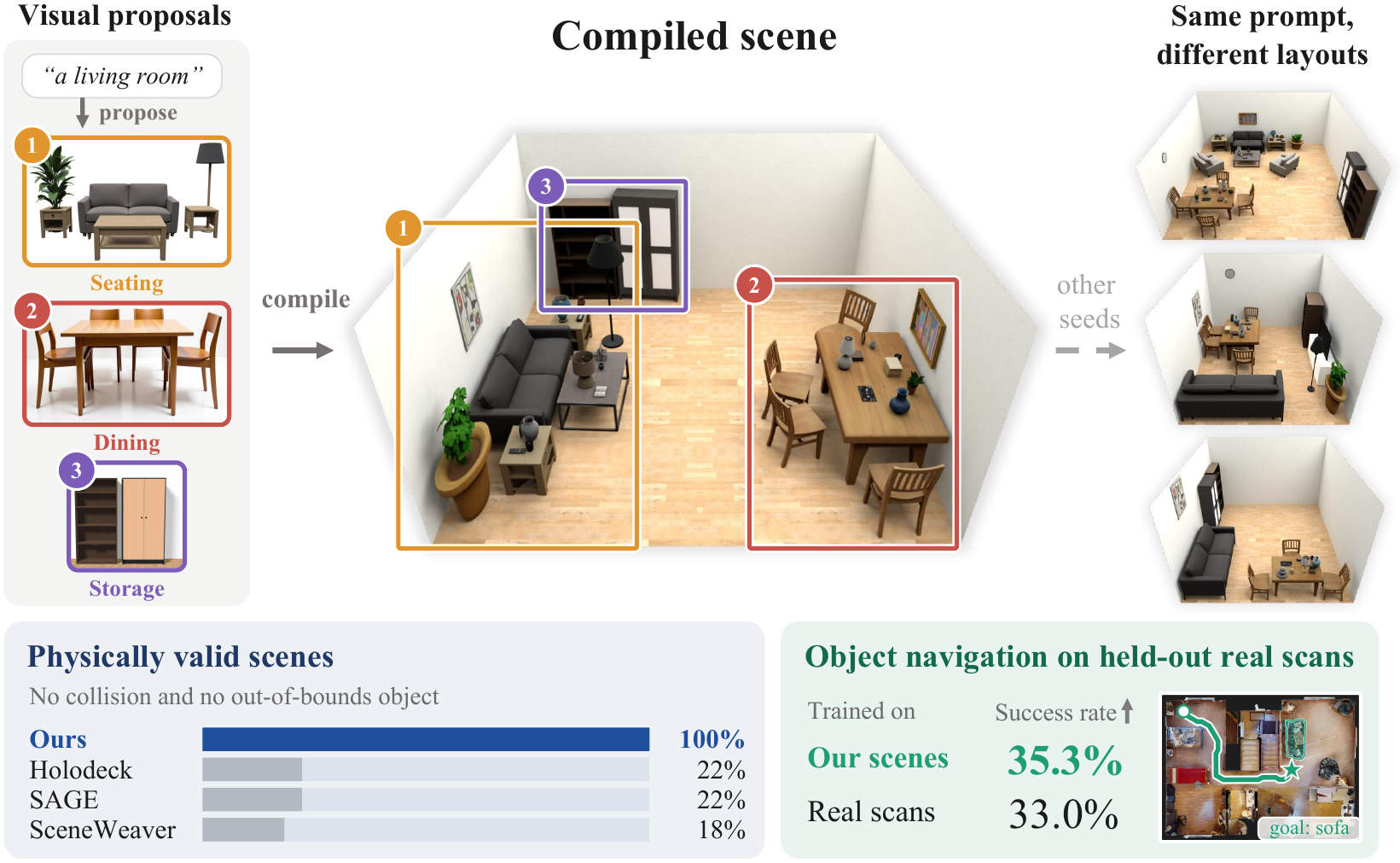}
\captionof{figure}{\textbf{Overview of SceneBloom.} SceneBloom compiles visual proposals of functional
zones into a simulation-ready 3D scene, and resampling the same prompt yields different layouts. Its scenes are more physically plausible than those of prior generators and train a better object-navigation policy than an equal number of real scans.}
\label{fig:teaser}
\end{center}

\input{sections/1_introduction}

\input{sections/2_related_work}

\input{sections/3_method}

\input{sections/4_experiment}

\input{sections/5_conclusion}

\bibliography{references}
\bibliographystyle{preprint}

\clearpage
\appendix
\input{sections/Appendix}

\end{document}

%% file: math_commands.tex
\usepackage{amsmath,amsfonts,bm}

\def\eqref#1{equation~\ref{#1}}

\def\1{\bm{1}}

\DeclareMathAlphabet{\mathsfit}{\encodingdefault}{\sfdefault}{m}{sl}
\SetMathAlphabet{\mathsfit}{bold}{\encodingdefault}{\sfdefault}{bx}{n}

%% file: required.tex
\usepackage{booktabs}        %
\usepackage{graphicx}        %
\usepackage{multirow}
\usepackage{colortbl}        %
\usepackage{tabularx}        %
\newcolumntype{Y}{>{\centering\arraybackslash}X}
\definecolor{oursbg}{RGB}{232,234,246}   %
\usepackage{amsmath,amssymb} %

\usepackage{xcolor}          %
\usepackage{enumitem}        %
\usepackage[capitalize,noabbrev]{cleveref}   %

\ifdefined\ifshowplan\else\newif\ifshowplan\fi
\showplanfalse
\providecolor{plancolor}{RGB}{110,60,10}
\providecolor{qcolor}{RGB}{20,90,140}
\providecolor{warncolor}{RGB}{160,20,20}

\providecommand{\budget}[2]{\ifshowplan\par\noindent{\color{plancolor}\sffamily\small\textbf{Budget:} #1 pages / #2 pages (compact)}\par\fi}

%% file: sections/1_introduction.tex
\section{Introduction}
\label{sec:intro}

Learning-based embodied agents benefit from training across varied
environments. Prior work shows that procedurally generated environments
support generalization across embodied tasks, and that language-guided
scene generation can improve navigation in new scene
types~\citep{procthor2022,holodeck2024}. 
These findings motivate generators
that provide diverse indoor layouts while retaining the spatial
organization of real environments.

Existing generators obtain layout structure from hand-crafted placement rules
or learned distributions of 3D
configurations~\citep{procthor2022,diffuscene2024,physcene2024}, while
language-guided and agentic systems combine flexible scene specification
with spatial reasoning and iterative
refinement~\citep{holodeck2024,sceneweaver2025,sage2026,scenesmith2026}.
Image diffusion models offer a complementary prior over how objects
are arranged together. Recent methods exploit generated images and
visual guidance to construct realistic 3D
scenes~\citep{architect2024,artiscene2025,scenethesis2025}. This
suggests a division of responsibilities in which semantic planning
establishes functional organization and visual sampling proposes
detailed arrangements. The challenge is to carry those arrangements
into geometrically valid 3D scenes.

Visual plausibility alone is insufficient for this realization. Depth, scale, and pose errors from 3D lifting produce overlapping or misaligned furniture when the reconstructed objects are assembled, so geometric correction is necessary. Moving or rotating objects to resolve these conflicts, however, can break the relationships that made the sampled arrangement plausible, for example by turning a chair away from its desk. Our guiding insight is to \emph{separate the
qualitative relationships to preserve from the numerical poses that
must be repaired}.

We introduce \textbf{SceneBloom}, an agentic framework that implements this separation through two stages, \emph{hierarchical visual sampling} and \emph{contract-guided 3D compilation}. In the first stage, a vision--language planner divides the room into functional zones, using retrieved exemplar rooms as references, and specifies each zone's furniture inventory and approximate floor region. A diffusion model then samples a visual proposal of each zone's furniture arrangement, so that zoning supplies room-level structure while each proposal models one furniture group jointly. The diffusion model uses asset-specific adapters, each trained on renders of one catalog asset, so that the proposals depict the objects used for 3D construction. In the second stage, the proposals are lifted into 3D. Before any object is moved, a vision--language model extracts \emph{relation contracts} from the proposal images, recording relationships such as facing, alignment, and proximity among the lifted objects. A hierarchical compiler then fits furniture within zones, assembles the zones into a room, and refines object poses to restore the contract relations. The contracts thus fix the relationships to preserve, while the compiler repairs only the numerical poses.

Across five room types, SceneBloom scores highest on every visual-quality
axis and on physical stability. It is also the only method with no detected mesh collisions and out-of-bounds objects. Its layouts more closely match real indoor
layout distributions than all compared baselines on both Matterport3D~\citep{matterport3d2017} and
Scan2CAD~\citep{avetisyan2019scan2cad}. Ablations show that visual proposals outperform textual proposals on every metric, and that contract guidance further improves layout coherence and realism. For object navigation, a policy trained from scratch in our houses achieves higher success on held-out real scans than one trained on an equal number of real scans from the same dataset. Randomizing object placement within the same rooms removes most of this advantage, indicating that it arises from layout organization. Fine-tuning on our scenes also improves a pretrained policy, and controllable generation yields further gains by matching the difficulty of real homes.

Our key contributions are:
\begin{itemize}[leftmargin=1.5em]
  \item We introduce an agentic framework for generating diverse,
    realistic, simulation-ready indoor environments for embodied
    learning. It combines diffusion-based layout sampling with
    contract-guided 3D compilation to preserve functional
    organization while resolving geometric conflicts.
  \item We show that our scenes are both visually realistic and
    physically valid, and that their layouts align more closely
    with real rooms than those of all compared generators.
  \item We demonstrate that object navigation policies trained on our
    scenes transfer to real scanned
    environments better than those trained on real scans, and that
    controllable scene difficulty strengthens this transfer.
\end{itemize}

%% file: sections/2_related_work.tex
\section{Related Work}
\label{sec:related}
\budget{0.55}{0.45}

\paragraph{3D Indoor Scene Synthesis}
Procedural methods assemble indoor scenes under hand-crafted spatial rules~\citep{procthor2022,infinigenindoors2024}, while autoregressive and diffusion models learn object configurations from annotated 3D scenes~\citep{atiss2021,diffuscene2024,instructscene2024}.
Physically guided variants further incorporate geometric constraints into layout sampling~\citep{physcene2024,spread2026}.
The arrangements of all these approaches remain bounded by the rules they encode or the 3D scenes they learn from.
To draw on broader visual knowledge, image-guided methods generate scene proposals through whole-scene synthesis or progressive inpainting, then recover and refine 3D object placements~\citep{artiscene2025,architect2024,scenethesis2025}.
Although these methods bring rich visual priors into scene generation, image-level plausibility does not ensure compatibility with actual asset geometry, and correcting the resulting conflicts can disrupt the relationships that make the proposed layout functional.
Our approach addresses both problems, since its proposals depict the actual assets to be placed and its compiler repairs their geometry under relation contracts extracted from the proposals.

\paragraph{Language-Guided and Agentic Scene Generation}
Language-guided methods prompt an LLM to write an object inventory and spatial constraints, which a solver satisfies~\citep{holodeck2024}.
Agentic systems turn generation into a feedback loop, in which a planner selects generation and editing tools step by step~\citep{sceneweaver2025,sage2026} or specialized agents divide scene design, critique, and workflow control~\citep{scenesmith2026,scenemosaic2026}.
Rendered views, geometric checks, and physics simulation guide each revision.
Other work structures layout decisions through functional-zone graphs, semantic and physical decomposition, or compilable layout programs~\citep{zonemaestro2026,stable2026,spatialgrammar2026}.
Iterative refinement, however, repeats tool calls, rendering, and review, which slows generation~\citep{sceneorchestra2026}, and errors in object orientation or in coordination across stages can leave arrangements functionally inconsistent~\citep{sceneweaver2025,scenesmith2026}.
Our framework uses agentic planning for functional zoning and diffusion sampling for furniture arrangement.
Each zone's arrangement is thus sampled in one pass rather than revised object by object, and relation contracts keep it functional during 3D compilation.

\paragraph{Synthetic Environments for Embodied Learning}
Synthetic environments provide controllable variation in the scenes and interactions for policy learning.
Procedural and language-guided generators expand training coverage through alternative layouts, scene types, and object categories~\citep{procthor2022,holodeck2024,divscene2024}.
Studies of synthetic navigation datasets further show that scene realism matters alongside dataset size, motivating attention to object occurrence and spatial arrangement patterns~\citep{hssd2024}.
For manipulation, simulation frameworks pair varied environments and tasks with demonstrations~\citep{robogen2023,robocasa2024,sage2026,robocasa3652026}.
Real-to-sim approaches construct training scenes from observations, including families of alternative assets and configurations that retain relevant geometric and semantic affordances~\citep{digitalcousins2024,simfoundry2026}.
Together, these directions motivate training environments that combine variation with realistic spatial organization.
SceneBloom provides both, and its generation settings can also adjust scene difficulty for policy learning.

%% file: sections/3_method.tex
\section{Method}
\label{sec:method}
\label{sec:formulation}

\begin{figure}[t]
    \centering
    \includegraphics[width=\linewidth]{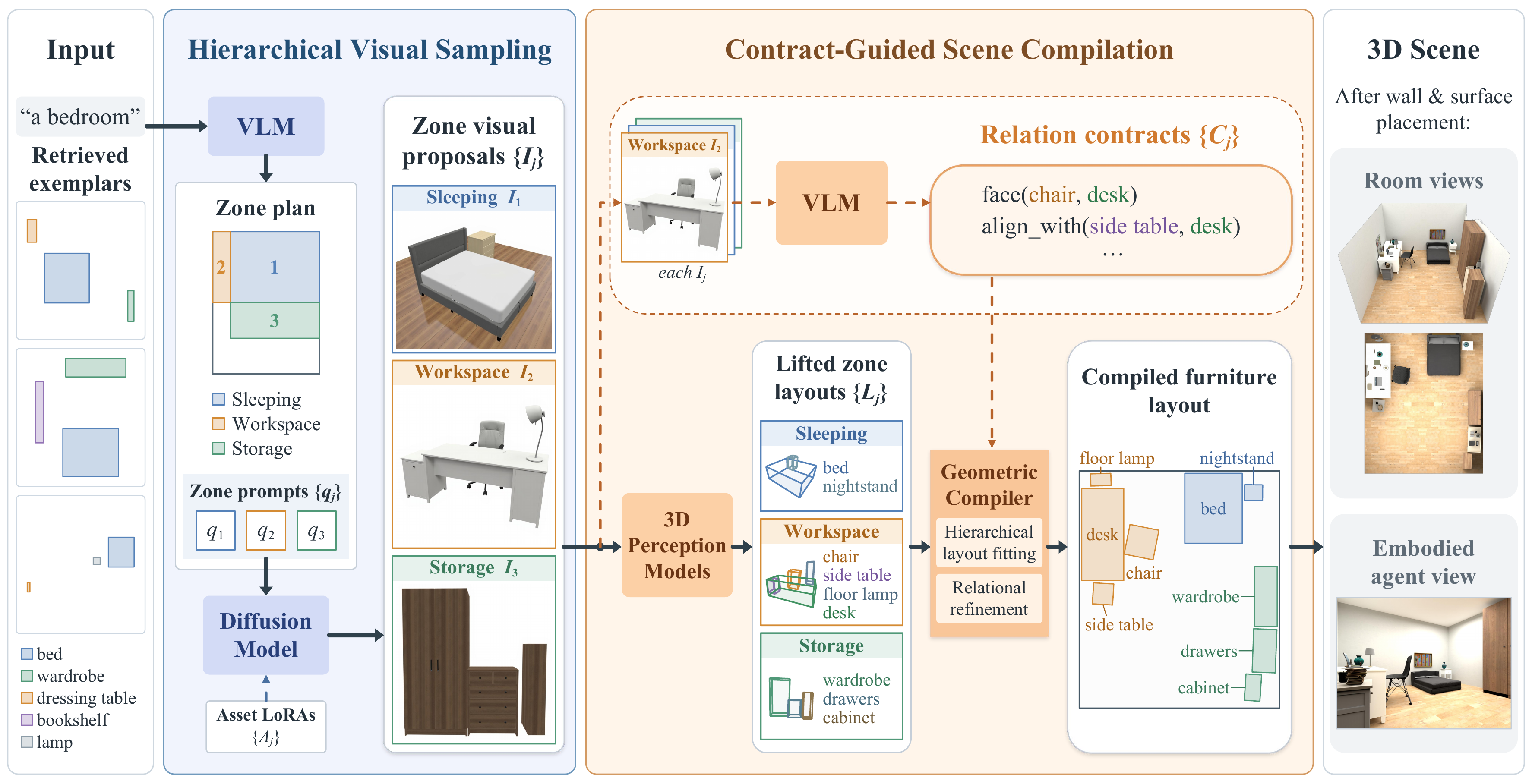}
    \caption{
         \textbf{Framework overview.}
        SceneBloom decouples visual layout sampling from 3D scene compilation.
        Given a room description and retrieved exemplars, a vision--language
        model (VLM) produces a zone plan with a prompt $q_j$ for each zone, and
        a diffusion model conditioned on $q_j$ and asset LoRAs $\Lambda_j$
        samples zone visual proposals $I_j$.
        3D perception models lift the proposals into zone layouts $L_j$, and a VLM extracts relation contracts $\mathcal{C}_j$ from the proposal images.
        Guided by these relation contracts, the geometric compiler applies
        hierarchical layout fitting and relational refinement to produce the
        compiled furniture layout.
        Wall and surface objects are then added, yielding the simulation-ready
        scene shown in room and agent views.}

    \label{fig:pipeline}
    \vspace{-0.1cm}
\end{figure}

Given a room description, SceneBloom generates a simulation-ready 3D scene in two stages (\cref{fig:pipeline}). Hierarchical visual sampling (\cref{sec:proposals}) divides the room into functional zones and samples a visual proposal of each zone's furniture arrangement. Contract-guided 3D compilation (\cref{sec:compiler}) lifts these proposals into 3D, fits them into the room, and restores their functional relations before adding wall and surface objects. This division lets the visual prior supply plausible arrangements while compilation makes them physically valid.

\subsection{Hierarchical Visual Sampling}
\label{sec:proposals}
Hierarchical visual sampling divides the work between room-level planning and zone-level arrangement. A vision--language model (VLM) plans the functional zones of the room, and a diffusion model then proposes the furniture arrangement of each zone as an image. Each proposal thus depicts one coherent furniture group, while the zone plan provides room-level structure.

\paragraph{Functional zoning.}
Deciding which furniture a room needs and how the room divides into functional zones requires commonsense knowledge that a VLM has. The VLM therefore determines the room dimensions and divides the room into functional zones $\{z_j\}_{j=1}^{J}$, using retrieved exemplar rooms as references (Appendix~\ref{app:proposal-details}). Each zone $z_j$ specifies a plan region, a furniture inventory $F_j$ with categories and counts, an anchor category for its central piece, the wall it adjoins, and a zone prompt $q_j$. Placement within each zone is left to visual proposals and geometric compilation.

\paragraph{Visual layout proposals.}
Image diffusion models learn from large image collections how furniture is typically arranged. We use this prior to propose the arrangement of each zone. Because the model generates an entire furniture group jointly, a proposal determines all relative placements at once rather than through sequential per-object decisions. For zone $z_j$, a text-to-image diffusion model $G$ is conditioned on the zone prompt $q_j$ and on asset LoRA adapters $\Lambda_j$ selected according to $F_j$:
\begin{equation}
I_j = G(q_j,\Lambda_j;\epsilon_j),
\end{equation}
where $I_j$ is the proposal image and $\epsilon_j$ is the diffusion seed. Each adapter is trained on renders of one catalog asset~\citep{ruiz2023dreambooth}, so that the proposal depicts the assets used for 3D construction. Adapter fusion~\citep{freefuse2025} allows several of these assets to appear in one image without interfering. 
Varying the seed $\epsilon_j$ then yields diverse arrangements of the same assets, each grounded in the objects to be built. Appendix~\ref{app:proposal-details} describes adapter training and fusion.

\subsection{Contract-Guided 3D Compilation}
\label{sec:compiler}

\begin{figure}[t]
    \centering
    \includegraphics[width=\linewidth]{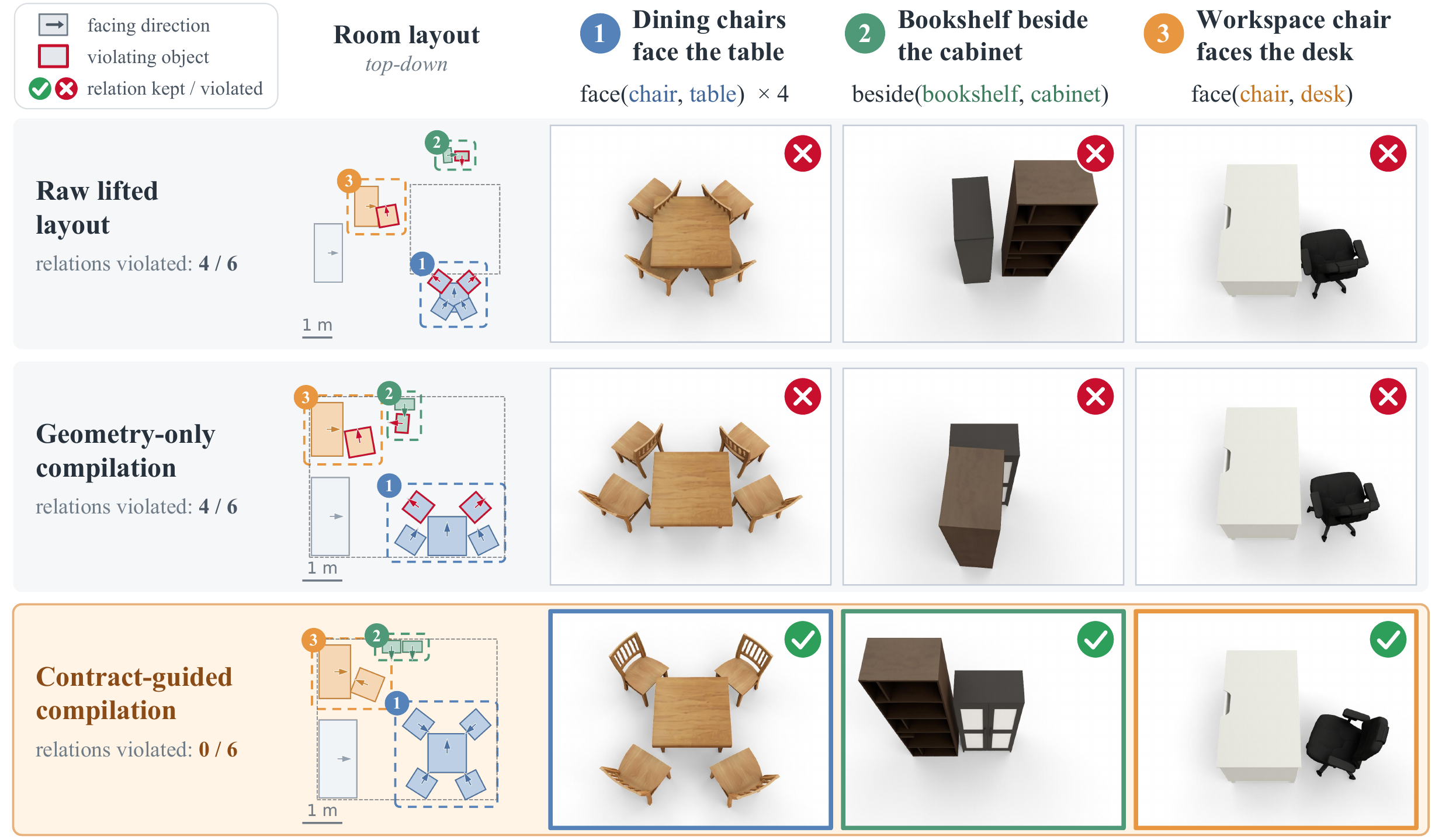}
    \caption{
        \textbf{Effect of contract-guided compilation.}
        All rows start from the same lifted scene, and each adds one step;
        the enlarged views show relation groups from the contract.
        \emph{Raw lifted layout}: objects keep their lifted poses, and lifting
        errors break relations such as chairs facing their table.
        \emph{Geometry-only compilation}: fitting assembles a collision-free
        room but leaves these relations broken.
        \emph{Contract-guided compilation}: fitting under the contract, followed
        by relational refinement, restores these relations while the layout
        stays collision-free.
    }
    \label{fig:contract-mechanism}
    \vspace{-0.1cm}
\end{figure}
Lifting a proposal into 3D introduces two kinds of error: geometric conflicts between objects and disrupted functional relations. Hierarchical layout fitting resolves the former, and relation contracts guide the recovery of the latter.

\paragraph{Lifting and relation extraction.}
\label{sec:lifting}
Monocular 3D perception~\citep{ovmono3d2024}, with depth and camera intrinsics from a monocular estimator~\citep{depthpro2024}, estimates a box for each object in $I_j$ and yields a zone layout $L_j$ with each object's category, dimensions, position, and orientation. To recover relations that geometric separation cannot restore, such as a nightstand beside its bed, the VLM extracts a relation contract $\mathcal{C}_j=\{(u,r,v,\gamma)\}$ from the proposal image and grounds it to the objects of $L_j$. In each entry, object $u$ holds relation $r$ to a target object or wall $v$ with confidence $\gamma$. The contract is fixed before compilation. Unlike language-authored constraints that prescribe a layout~\citep{holodeck2024}, it records relations present in the sampled arrangement. Its high-confidence relations ($\gamma \ge 0.75$) guide the rest of compilation (Appendix~\ref{app:contract-details}).

\paragraph{Hierarchical layout fitting.}

We resolve geometry within zones and then assemble the zones according to the zone plan, reconciling the planned zone regions with the footprint that the proposed furniture actually needs.
\emph{Within each zone}, the anchor of $z_j$ is aligned with the zone's wall and faces into the room. The remaining objects of $L_j$ are then separated from the fixed anchor and from one another to remove footprint overlaps, with $\mathcal{C}_j$ guiding how related objects are separated. The fitted objects define the zone envelope $E_j$, which is then frozen so that room-level assembly moves the zone only as a rigid whole.
\emph{Among zones}, the zone plan fixes for each pair of zones $(z_i,z_j)$ a separating axis and their order along it. Along that axis, an envelope bound to a wall stays flush against it, any other envelope keeps its planned position where possible, and otherwise assembly shifts the later envelope along the axis until it starts at least a clearance gap after the earlier one ends. The room is enlarged only when the envelopes cannot fit within it.
Fitting thus resolves geometric conflicts while preserving the arrangement of each proposal within its zone.

\paragraph{Relational refinement.}
After assembly, geometric predicates check each high-confidence relation of $\mathcal{C}_j$. Relation-specific operators propose local rotations or translations to restore violated relations, subject to room-boundary and footprint feasibility checks. Because these adjustments are local, they restore relations without disturbing the arrangement sampled in the proposal. Relations such as chairs facing their table decide whether a zone is functional and coherent, so restoring them directly supports visual quality. \Cref{fig:contract-mechanism} contrasts geometry-only compilation, which omits the contract, with contract-guided compilation.

\paragraph{Scene completion.}
Finally, the VLM adds wall-mounted objects and small objects on furniture surfaces. For each wall or furniture top, it chooses object categories and suggests positions in that surface's coordinates, and geometric placement maps these positions into the room while checking available space, boundaries, and intersections (Appendix~\ref{app:completion}).

%% file: sections/4_experiment.tex
\section{Experiments}
\label{sec:eval}

\subsection{Scene Generation}
\label{sec:generation}

\paragraph{Implementation Details.}
We use Qwen3-VL-8B-Instruct~\citep{bai2025qwen3} for functional zoning and scene compilation, and FLUX.1-dev~\citep{flux2024} with FreeFuse~\citep{freefuse2025} and asset-specific LoRA adapters for visual layout proposals.
OVMono3D-LIFT~\citep{ovmono3d2024} lifts 2D boxes from Grounding DINO~\citep{liu2023grounding} into 3D, with depth and camera intrinsics from Depth Pro~\citep{depthpro2024}.
Reference exemplars are drawn from M3DLayout~\citep{m3dlayout2026}.
Appendix~\ref{app:generation} details generation configurations, asset sources, and reference-data separation.

\paragraph{Baselines.}
We compare against three representative methods using their official codebases.
Holodeck~\citep{holodeck2024} uses an LLM to specify spatial constraints for a layout solver.
SceneWeaver~\citep{sceneweaver2025} uses a self-reflective agent to coordinate generation and editing tools.
SAGE~\citep{sage2026} uses visual and physics critics to refine scenes and validate placements in simulation.
We retain method-specific asset sources, use our Qwen3-VL-8B-Instruct backbone for Holodeck and SceneWeaver for a controlled comparison, and keep the model pair specified in SAGE's paper.
Appendix~\ref{app:generation} details these adaptations.

\paragraph{Metrics.}
We evaluate physical plausibility, visual quality, and real-layout alignment.
Following SceneEval~\citep{sceneeval2026}, we report mesh-level collision rates at the object (Coll.\ obj) and scene (Coll.\ scene) levels and mean out-of-bounds objects per scene (OOB).
We also report the stable-object rate (Stab.) under REST3D~\citep{rest3d2026}.
Following SceneWeaver and SAGE~\citep{sceneweaver2025,sage2026}, GPT-4o~\citep{hurst2024gpt} scores realism (Real.), functionality (Func.), layout coherence (Coh.), and completeness (Compl.).
For real-layout alignment, \emph{Layout Distribution Discrepancy} (LDD) applies unbiased RBF maximum mean discrepancy (MMD$^2$)~\citep{gretton2012mmd} to features encoding object composition and spatial relations, against two references of real rooms: Matterport3D~\citep{matterport3d2017} and Scan2CAD~\citep{avetisyan2019scan2cad}.
Appendices~\ref{app:metrics} and~\ref{app:results} provide metric definitions, scoring protocols, hardware, and resource accounting.

\subsubsection{Main Results}
\label{sec:generation-results}
We generate and evaluate 10 scenes for each of five room types: bedrooms, living rooms, dining rooms, bathrooms, and offices, using the template \texttt{a \textless room type\textgreater}.

\paragraph{Physical Plausibility.}
Our scenes are the only ones free of mesh collisions and out-of-bounds objects, and they are the most stable under REST3D (\cref{tab:quality}).
The baselines add objects one at a time without resolving the scene as a whole, so conflicts accumulate as a scene grows: SAGE's agent uses 407 of its 504 tool calls across its 50 scenes for placement but moves a placed object in only 44, and Holodeck spawns repeated copies of an asset at a single point.
The instability also reaches the furniture itself: $31\%$ of Holodeck's and $26\%$ of SAGE's and SceneWeaver's large objects, those at least $0.6$\,m across, move beyond the stability threshold, against $3\%$ of ours.
Our compiler instead resolves all furniture of a room jointly before completion and verifies each surface object against its support, so no conflict is left to accumulate. Additional stability evaluations appear in Appendix~\ref{app:results}.

\begin{figure}[t]
    \centering
    \includegraphics[width=\linewidth]{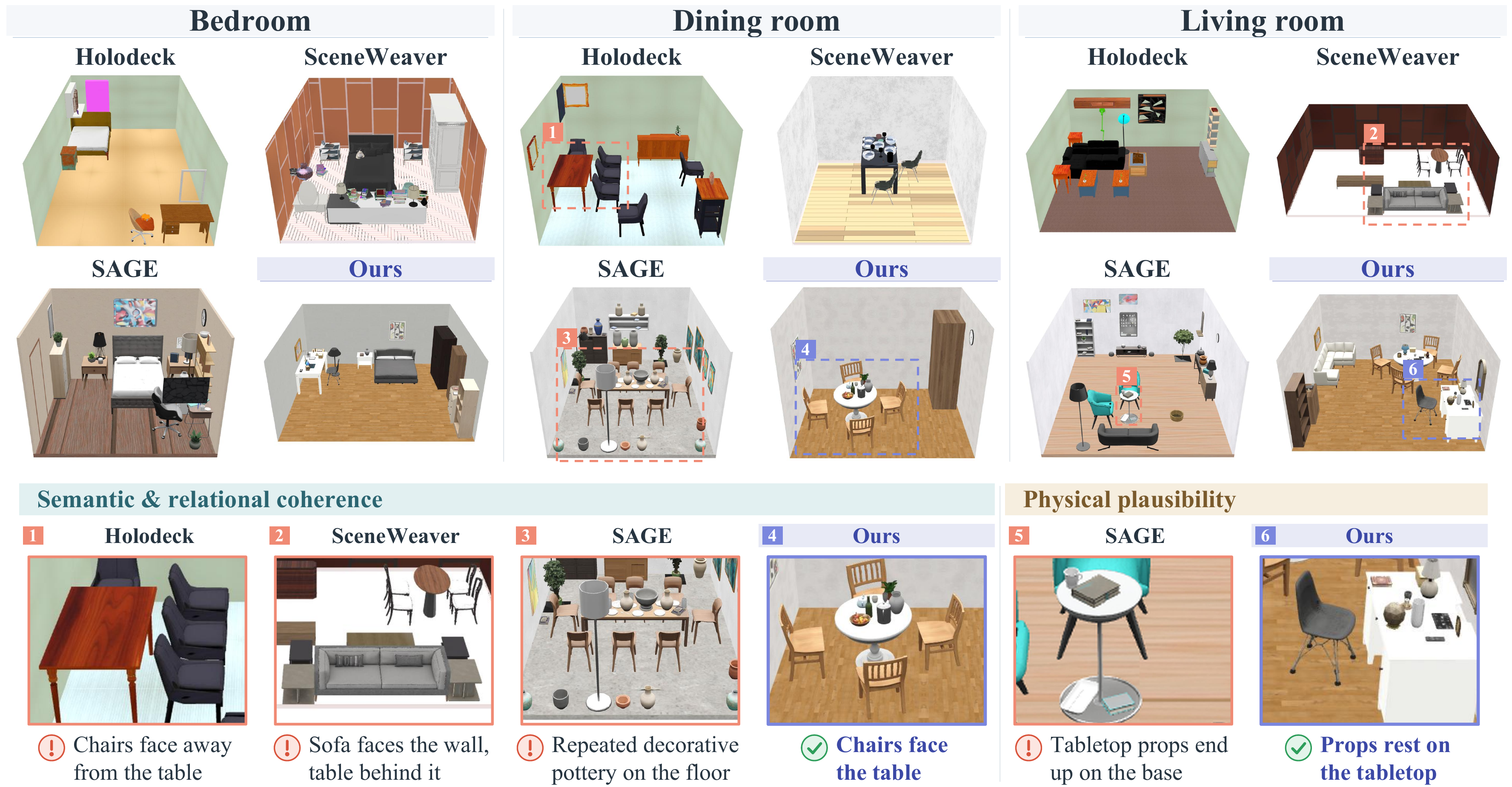}
    \caption{
        \textbf{Qualitative comparison.}
        For each method and room type, we show the scene whose score is the median of its ten scenes.
        Enlarged regions illustrate functional organization
        and physical placement across methods, each retaining its native asset source.
        Aggregate results are in \cref{tab:quality}.
    }
    \label{fig:qualitative}
\end{figure}

\paragraph{Visual Quality.}
Our scenes score highest on all four visual-quality axes.
SAGE is a strong baseline and places the most objects, but its placement loop keeps adding them until floors and surfaces are crowded (\cref{fig:qualitative}), while our scenes score higher on functionality and completeness with fewer and reasonable objects.
Holodeck and SceneWeaver place furniture through textual constraints and edits, so functional relations are often lost, for example, dining chairs that stand apart from the table and face away from it.
Our proposals depict each functional group as a whole, and the contract preserves its relations through compilation. Section~\ref{sec:ablation} further isolates the effect of arrangement by holding the inventory fixed. 

\paragraph{Real-Layout Alignment.}
Our layouts are the closest to both real-room references, Matterport3D and Scan2CAD.
Because LDD compares which objects a room contains and how they are arranged around its anchor furniture, a low distance on both references indicates that our rooms are organized like real rooms rather than fitted to a single dataset.
Held-out real rooms score near zero against their own reference, far below every method, so these distances reflect genuine differences in layout organization (Appendix~\ref{app:distribution}).

\paragraph{Efficiency.}
Generation takes 2.7 minutes per layout, approximately $27\times$ faster than SAGE on the same hardware, because each zone's arrangement is sampled once and compiled without an iterative placement loop. Holodeck is faster per layout (Appendix~\ref{app:results}), but $78\%$ of its scenes contain collisions.

\input{tables/scene_quality_main}

\subsubsection{Ablation Study}
\label{sec:ablation}
To isolate visual sampling and contract guidance, we replace the visual proposal with a textual proposal, in which a VLM directly specifies the position and orientation of each object.
Each proposal is compiled with and without the contract extracted from it (\cref{tab:proposal-compiler-ablation}).

Visual proposals provide a stronger foundation for compilation, and contract guidance further improves them while preserving physical plausibility.
Under both contract settings, visual proposals outperform textual proposals on every physical and visual-quality metric.
Holding the visual proposal fixed, contract guidance additionally raises layout coherence from $4.31$ to $4.65$ and realism from $5.66$ to $5.83$, consistent with its role in restoring object poses and relations.
Contract guidance also cuts contract violations from $37.45\%$ to $4.25\%$ (Appendix~\ref{app:ablations}), and the compiled layouts remain collision-free.
Visual sampling thus establishes layout quality, and contract guidance preserves the sampled relations through 3D compilation.

For textual proposals, contract guidance improves visual quality but increases collisions and reduces stability.
These proposals are less plausible to begin with, as their objects are more crowded than in visual proposals and some already collide.
Restoring relations within such arrangements can introduce further conflicts.
Contract guidance therefore refines a plausible proposal but cannot substitute for one, which is why SceneBloom pairs it with visual sampling.

\begin{figure}[t]
    \centering
    \includegraphics[width=\linewidth]{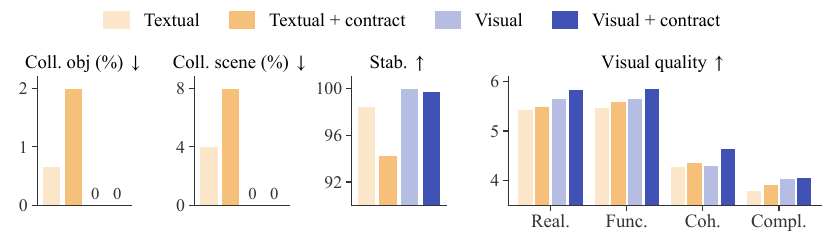}
    \caption{
        \textbf{Ablation of proposal source and contract guidance.}
        Results on 50 furniture-only scenes with shared inventories, assets, and sizes.
        Appendix~\ref{app:ablations} lists the exact values.
    }
    \label{fig:proposal-compiler-ablation}
    \vspace{-0.2cm}
\end{figure}

\subsection{Object-Goal Navigation Learning}
\label{sec:learning}
We use our scenes to train object-goal navigation policies in two settings.
A modular policy is trained from scratch, so the scenes provide all of its experience (\cref{sec:scratch}), and a pretrained open-vocabulary policy is fine-tuned on them (\cref{sec:finetune}).

\paragraph{Training Sources and Evaluation.}
We import scenes into Habitat~\citep{savva2019habitat} and construct episodes from visible goal viewpoints and reachable starts.
Both settings compare SceneBloom (ours) with Randomized (the same scenes with floor-object poses resampled), Agent-planned (text-agent layouts with zoning), and real scans from HM3D~\citep{ramakrishnan2021hm3d}, and train every source with an identical recipe and budget.
We evaluate in held-out real scanned environments and report success rate (SR) and success weighted by path length (SPL)~\citep{anderson2018evaluation}.
Appendix~\ref{app:training} gives further training details.

\subsubsection{Training from Scratch}
\label{sec:scratch}

\begin{wrapfigure}[18]{r}{0.48\linewidth}
  \centering
  \vspace{-12pt}
  \includegraphics[width=\linewidth]{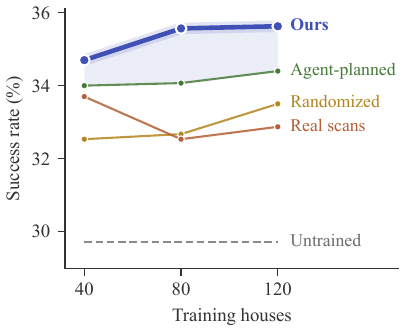}
  \caption{\textbf{Training from scratch.} SR (\%) on the 36 real HM3D validation homes. Points average three training seeds. Shading marks our margin over the best other source.}
  \label{fig:scratch}
\end{wrapfigure}
\paragraph{Setup.}
SemExp~\citep{chaplot2020semantic} builds a top-down semantic map and uses a learned goal policy to choose where to search.
We train this goal policy from scratch with PPO~\citep{schulman2017ppo} for 2M steps in the framework of \citet{wang2026whatmatters}.
Each source provides 40, 80, or 120 houses, and each of our houses connects three to five generated rooms along a corridor.
We evaluate on 1,000 episodes in 36 real HM3D validation homes, where success follows the SemExp setup and requires calling STOP within 1\,m of a goal object (Appendix~\ref{app:scratch}).
At evaluation, the policy reads semantics predicted from RGB by a COCO-trained Mask R-CNN~\citep{he2017mask}.

\paragraph{Real-Environment Transfer.}
Policies trained on our houses transfer best to real scanned environments, outperforming those trained on real scans, Randomized houses, and Agent-planned houses at every house count (\cref{fig:scratch}).
Our houses raise SR from 29.7\% for the untrained policy to 35.3\% on average, $1.7\times$ the gain from training on real scans, even though the real scans come from the same dataset as the evaluation homes.
Our houses also give the policy more complete searches to learn from.
Their training episodes end at the policy's own STOP in 91.9\% of cases, against 75.0\% in real scans, and the gap holds at every start distance (Appendix~\ref{app:scratch}).
Our generator therefore supplies more effective training experience than real scans and scales more easily.

Randomized houses place the same furniture in the same rooms at random, which often splits a room's walkable floor and confines training searches to part of the house, whereas our arrangements rarely split it (Appendix~\ref{app:navmesh}).
Agent-planned houses share our functional zones and assets, but a text agent places objects less realistically than our visual proposals (\cref{sec:ablation}), so the policy learns to search among arrangements that are less common in real homes.

\subsubsection{Fine-Tuning a Pretrained Policy}
\label{sec:finetune}

\paragraph{Setup.}
We fine-tune OVSegDT~\citep{ovsegdt2025}, an open-vocabulary policy conditioned on RGB observations, language goals, target masks, and history.
Every policy is fine-tuned from the released OVSegDT checkpoint (the pretrained policy) with PPO for 128,000 environment steps under an identical recipe.
Each generated source provides 120 scenes, 24 for each of the five room types in \cref{sec:generation}, and real-scan continuation applies the same recipe to 142 HM3D training scans.
We evaluate on held-out scenes from our generator and an 864-episode subset of HM3D-OVON Val Unseen~\citep{hm3dovon2024} covering all 36 of its scenes, where success requires calling STOP action within 1.0\,m of the goal.

\paragraph{Generalization to Unseen Goals and Environments.}
\label{sec:utility}
Fine-tuning on our scenes achieves the highest success rates on unseen goals in both held-out generated and real scanned environments (\cref{tab:learning}).
For goals used during fine-tuning, it raises SR from 73.67\% to 74.60\% and SPL from 61.31\% to 64.00\%, and Randomized and Agent-planned layouts yield similar gains.
Further, on goal categories not used during fine-tuning, our scenes yield the highest SR and SPL, whereas every other fine-tuned policy falls below the pretrained policy in SR.
In real scanned environments, where both the scenes and the goal categories are unseen, our scenes also outperform real-scan continuation, which only revisits the 142 HM3D scans seen during pretraining.
Our generator instead supplies new and varied rooms on demand, so fine-tuning on 120 generated scenes generalizes better than on 142 real scans.

\begin{table}[h]
  \centering
  \caption{\textbf{Policy fine-tuning and generalization.}
  SR and SPL (\%) on 100 seen-goal and 107 unseen-goal held-out generated scenes and on the
  36 scanned scenes of the HM3D-OVON subset. Fine-tuned rows average three training seeds.
  Bold/underline mark best/second-best.}
  \label{tab:learning}
  \small
  \begin{tabularx}{\linewidth}{l*{6}{Y}}
  \toprule
  \multirow{3}{*}{Policy} & \multicolumn{4}{c}{Held-out generated environments} & \multicolumn{2}{c}{Real environments} \\
  \cmidrule(lr){2-5}\cmidrule(l){6-7}
  & \multicolumn{2}{c}{Seen goals} & \multicolumn{2}{c}{Unseen goals} & \multicolumn{2}{c}{Unseen goals} \\
  \cmidrule(lr){2-3}\cmidrule(lr){4-5}\cmidrule(l){6-7}
  & SR & SPL & SR & SPL & SR & SPL \\
  \midrule
  Pretrained (no fine-tuning) & 73.67 & 61.31 & \underline{52.57} & 38.29 & \underline{48.26} & \textbf{21.18} \\
  \midrule
  \multicolumn{7}{l}{\textit{Fine-tuned on 142 real scans}} \\
  Real-scan continuation & 73.02 & 61.12 & 51.59 & \underline{38.42} & 47.57 & 20.68 \\
  \cmidrule(lr){1-7}
  \multicolumn{7}{l}{\textit{Fine-tuned on 120 generated scenes}} \\
  Randomized & \textbf{74.88} & 63.26 & 51.51 & 38.09 & 48.15 & 20.74 \\
  Agent-planned & 74.39 & \underline{63.39} & 51.63 & 37.65 & 47.99 & 20.72 \\
  \rowcolor{oursbg}
  SceneBloom (ours) & \underline{74.60} & \textbf{64.00} & \textbf{53.19} & \textbf{38.89} & \textbf{48.57} & \underline{21.17} \\
  \bottomrule
  \end{tabularx}
\end{table}

\paragraph{Controllable Scene Difficulty.}
\label{sec:occlusion}
Matching the difficulty of real scanned environments further improves transfer (\cref{tab:occlusion}).
Real multi-room homes often hide the goal behind walls and furniture, whereas our single rooms keep more goals in view.
Because SceneBloom organizes each room into functional zones, it can introduce this difficulty directly.
Using the same prompts as the 120 scenes in \cref{tab:learning}, we regenerate each scene with an additional interior zone of tall wardrobes (SceneBloom with occluders).
As a control, Randomized with occluders keeps these occluders fixed and resamples the other floor objects, and all policies are fine-tuned with the identical recipe.

\begin{wraptable}{r}{0.45\linewidth}
  \vspace{-2pt}
  \centering
  \caption{\textbf{Real-environment transfer with occluders.} SR and SPL (\%).}
  \label{tab:occlusion}
  \vspace{-4pt}
  \small
  \setlength{\tabcolsep}{4pt}
  \begin{tabularx}{\linewidth}{l*{2}{Y}}
  \toprule
  Policy & SR & SPL \\
  \midrule
  Pretrained & 48.26 & \underline{21.18} \\
  Real-scan continuation & 47.57 & 20.68 \\
  \midrule
  \multicolumn{3}{l}{\textit{Without occluders}} \\
  SceneBloom & \underline{48.57} & 21.17 \\
  \cmidrule(lr){1-3}
  \multicolumn{3}{l}{\textit{With occluders}} \\
  Randomized & 48.34 & 20.72 \\
  \rowcolor{oursbg}
  SceneBloom & \textbf{49.50} & \textbf{21.35} \\
  \bottomrule
  \end{tabularx}
  \vspace{-8pt}
\end{wraptable}
Three comparisons explain this gain.
First, the added difficulty itself helps, as success in real scanned environments rises from 48.57\% to 49.50\%.
Second, the difficulty is effective when paired with a coherent layout, as randomizing the other floor objects around the same occluders yields only 48.34\%.
Third, this generated variation is more useful than further training on real scans, as SceneBloom with occluders also surpasses real-scan continuation (47.57\%) under the same training budget.
Therefore, these results show how controllable generation supplies useful training variation by combining task-relevant difficulty with coherent spatial organization.
Appendix~\ref{app:finetune-evidence} quantifies the difficulty that each scene source introduces and validates these gains on real episodes cleaned of annotation errors.

%% file: tables/scene_quality_main.tex
\begin{table}[t]
\centering
\caption{\textbf{Comparison with scene generation methods.} Coll.\ and Stab.\ are percentages, OOB counts objects per scene. Bold/underline mark best/second-best.}
\label{tab:quality}
\small
\setlength{\tabcolsep}{3pt}
\resizebox{\linewidth}{!}{%
\begin{tabular}{l c cccc cccc cc c}
\toprule
\multirow{3}{*}{Method} & \multirow{3}{*}{\#Obj} & \multicolumn{4}{c}{Physical Plausibility} & \multicolumn{4}{c}{Visual Quality $\uparrow$} & \multicolumn{2}{c}{LDD $\downarrow$} & Time $\downarrow$ \\
\cmidrule(lr){3-6}\cmidrule(lr){7-10}\cmidrule(lr){11-12}\cmidrule(lr){13-13}
& & \multicolumn{2}{c}{Coll.\ $\downarrow$} & \multirow{2}{*}{OOB $\downarrow$} & \multirow{2}{*}{Stab.\ $\uparrow$} & \multirow{2}{*}{Real.} & \multirow{2}{*}{Func.} & \multirow{2}{*}{Coh.} & \multirow{2}{*}{Compl.} & \multirow{2}{*}{MP3D} & \multirow{2}{*}{Scan2CAD} & \multirow{2}{*}{(min)} \\
\cmidrule(lr){3-4}
& & obj & scene & & & & & & & & & \\
\midrule
Holodeck & 18.0 & 32.7 & \underline{78.0} & .160 & 35.1 & 4.66 & 5.07 & 4.23 & 4.30 & .2509 & .2342 & \textbf{1.6} \\
SceneWeaver & 18.0 & 47.2 & 82.0 & \textbf{.000} & \underline{45.3} & 5.30 & 5.41 & 4.30 & 5.06 & \underline{.2009} & \underline{.1797} & 313.0 \\
SAGE & 34.2 & \underline{12.7} & \underline{78.0} & \underline{.020} & 45.1 & \underline{5.91} & \underline{5.62} & \underline{4.54} & \underline{6.46} & .2191 & .2251 & 73.6 \\
\midrule
\rowcolor{oursbg}
SceneBloom (ours) & 23.6 & \textbf{0.0} & \textbf{0.0} & \textbf{.000} & \textbf{75.3} & \textbf{6.18} & \textbf{6.29} & \textbf{4.76} & \textbf{6.74} & \textbf{.1759} & \textbf{.1340} & \underline{2.7} \\
\bottomrule
\end{tabular}}
\end{table}

%% file: sections/5_conclusion.tex
\section{Conclusion}

\label{sec:conclusion}
We presented SceneBloom, an agentic framework that separates the relationships to preserve in a layout from the poses to repair. Visual proposals supply realistic arrangements, and contract-guided compilation repairs their geometry while keeping the proposed relations, so our scenes are physically valid, visually coherent, and closest to real room layouts. Sampling each zone once and resolving each room jointly also makes generation far faster than iterative agentic methods. These scenes in turn improve real-environment transfer for navigation policies trained from scratch and unseen-goal generalization after fine-tuning. Further gains from matching scene difficulty to real homes demonstrate the value of combining realistic layouts with adjustable task difficulty for embodied learning.

%% file: sections/Appendix.tex
\setcounter{topnumber}{3}
\section{Implementation Details}
\label{app:generation}

\subsection{Baseline Configurations}

We evaluate Holodeck~\citep{holodeck2024}, SceneWeaver~\citep{sceneweaver2025}, and SAGE~\citep{sage2026} from their released codebases, retaining their native asset sources and scene-generation procedures. Prompts, solver settings, and asset sources are unchanged; Holodeck and SceneWeaver run on Qwen3-VL-8B-Instruct, the backbone of our pipeline, and SAGE keeps the model pair specified in its paper, GPT-OSS-120B~\citep{agarwal2025gpt} with Qwen3-VL-30B-A3B-Instruct (\cref{tab:app-resource-cost}); the tool set follows the release's own instruction to enable the tools one can supply. Runtime is measured over each method's executable released pipeline, and \cref{tab:app-resource-time} states the included stages and hardware.

\paragraph{Asset library.}
Our asset library holds 192 assets: 75 floor-furniture meshes across 22 categories, 109 surface objects, and 8 wall objects. We generate most of them using the TRELLIS~\citep{xiang2025trellis} text-to-3D model. The 27 articulated floor meshes, which cover eight storage and sanitary categories, come from the ArtVIP dataset~\citep{jin2026artvip} instead. Our library is also extensible: once a category-compatible mesh has a trained LoRA adapter, the visual sampler can draw it, and the VLM, the compiler, and the prompts stay unchanged. Each baseline keeps its own asset source. Holodeck retrieves from Objaverse~\citep{deitke2023objaverse}, SAGE generates its assets with TRELLIS text-to-3D, and SceneWeaver generates most objects procedurally with Infinigen~\citep{infinigenindoors2024} and retrieves the rest from 3D-FUTURE~\citep{fu20213d} and Objaverse. Each of these systems treats its asset pipeline as part of its contribution, so substituting our meshes would evaluate a different system.

\subsection{Zone-Wise Visual Proposals}
\label{app:proposal-details}
\label{app:lora}
For each room, retrieval restricts M3DLayout~\citep{m3dlayout2026} exemplars to the requested room type and ranks them by cosine similarity between Qwen3-Embedding~\citep{zhang2025qwen3} encodings of the room description and of each exemplar's structured text. The top-ranked exemplar is kept as an anchor, and two more are drawn from the remaining candidates by similarity-weighted sampling without replacement. This varies the structural references while keeping the nearest neighbor.

The retrieval corpus is built from two disjoint indices. The first holds 5{,}173 3D-FRONT~\citep{fu20213dfront} rooms and covers bedrooms, living rooms, and dining rooms. The second holds 627 Matterport3D regions and covers bathrooms, offices, kitchens, lounges, recreation rooms, and meeting rooms; it contains no bedroom, living-room, or dining-room region. The real-layout references of \cref{app:layout-metrics} are Matterport3D bedrooms, living rooms, and dining rooms, which neither index contains, and Scan2CAD rooms, which come from ScanNet and are absent from both. No room that enters either reference distribution can therefore be retrieved as an exemplar, and the conditioning set and the evaluation references are disjoint by construction.

Every floor-furniture mesh the sampler can draw has its own LoRA adapter, trained on multi-view renders of that mesh with DreamBooth and prior preservation~\citep{ruiz2023dreambooth} and orthogonal adaptation~\citep{Po_2024_CVPR}; wall and surface objects are placed geometrically after compilation and need no adapter. The pipeline inserts each adapter's trigger token before the category word it stands for, so the word sofa in a zone prompt yields the particular sofa selected from our library. A zone uses at most five adapters, one per furniture category, and FreeFuse~\citep{freefuse2025} runs them together in a single generation, binding each adapter to the tokens that name its object and to the image region those tokens attend to. Every furniture object compiled from the image is therefore the library mesh whose adapter drew it. This grounding motivates the adapters: it reduces the appearance mismatch between a proposal and the objects used for construction, so the compiler rebuilds the sampled arrangement from the objects the image shows rather than from meshes that merely share their categories.

\subsection{Contract-Guided Scene Compilation}
\label{app:contract-details}
Before any geometry is fitted, the VLM reads each zone's original proposal image once. The VLM returns a fixed contract $\mathcal C_j=\{(u,r,v,\gamma)\}$ holding only the relations the proposal itself shows, where $u$ is a reconstructed object and $v$ is a reconstructed object or a wall, $r$ is one of the relations in \cref{tab:app-contract-relations}, and $\gamma$ is the VLM's confidence. Only relations with $\gamma \ge 0.75$ guide compilation. The prompt and output schema are given in Appendix~\ref{app:prompts}.

\input{tables/app_contract_relations}

The compiler then works outward. It first resolves each zone on its own, separating that zone's objects around its anchor under the guidance of the zone's contract, and records the zone's internal pairwise distances. It next places the zones within the room plan, moving every zone rigidly, and the recorded distances confirm that no zone is deformed. Finally, it checks each high-confidence relation and applies the rotation or translation that restores each violated one.

\subsection{Wall and Surface Object Placement}
\label{app:completion}

Wall objects are placed first and surface objects second, and both stages follow the same propose-then-verify pattern. The VLM chooses object categories, each realized by a seeded library mesh, and proposes positions in normalized coordinates of the target surface, along and up a wall or across a furniture top. Geometry converts these positions into room coordinates and rejects placements that collide with objects already placed. The two conversions follow.

For a wall with unit tangent $\mathbf t$, inward normal $\mathbf n$, and line point $\mathbf p_0$ at along-wall coordinate $0$, let $[a_0,a_1]$ be the admissible along-wall coordinates of the object's center within its zone. A wall object of depth $d$ with normalized proposal $(u,v)\in[-1,1]^2$ is mapped to
\begin{equation}
 a=a_0+\frac{u+1}{2}(a_1-a_0),\qquad
 \mathbf p_{xz}=\mathbf p_0+a\mathbf t+\frac{d}{2}\mathbf n,
\end{equation}
while $v$ selects a valid vertical position between the floor and ceiling. The object's yaw faces the wall normal. If the proposal refers to supporting furniture, its horizontal coordinate is aligned to the furniture center and its base height is placed above the furniture top. Candidates are rejected when their wall rectangles overlap or their 3D boxes intersect furniture.

For a support surface centered at $\mathbf c=(c_x,c_z)$ with half extents $(h_x,h_z)$, yaw $\theta$, and top height $y_{\mathrm{top}}$, the VLM proposes $(u,v,\psi)$, where $(u,v)\in[-1,1]^2$. The object's world pose is
\begin{equation}
 \mathbf p_{xz}=\mathbf c+R(\theta)
 \begin{bmatrix}u h_x\\v h_z\end{bmatrix},\qquad
 y=y_{\mathrm{top}}+\delta,\qquad
 \theta_{\mathrm{obj}}=\theta+\psi,
\end{equation}
where $\delta$ is a small clearance. We retain a placement only when its oriented footprint lies inside the inset support rectangle and does not overlap an accepted object.

\section{Evaluation Protocols and Metric Definitions}
\label{app:metrics}

\subsection{Layout Distribution Discrepancy}
\label{app:layout-metrics}

Layout Distribution Discrepancy (LDD) measures how closely generated layouts follow the distribution of real layouts. We use two independently collected references of real scanned rooms. Matterport3D~\citep{matterport3d2017} is read through the M3DLayout annotations, which express each room's furniture as oriented boxes, and Scan2CAD~\citep{avetisyan2019scan2cad} aligns CAD models to ScanNet~\citep{dai2017scannet} scans. For each room type, every annotated room of a reference is pooled into one empirical distribution with no model fitted to it (\cref{app:scan2cad} lists the counts).

We map generated and reference objects to a shared coarse taxonomy. For each room type, we define an anchor category $a$ and associated categories $\mathcal A$: bed with storage and table in bedrooms, table with chairs in dining rooms, and table with sofas and chairs in living rooms. Each associated object is assigned to its nearest anchor in the ground plane.

For anchor instance $j$ and category $c$, let $\Delta_{jck}=\mathbf p_{ck}-\mathbf p_{a_j}$, $r_{jck}=\lVert\Delta_{jck}\rVert_2$, and $\theta_{jck}=\operatorname{atan2}(\Delta^z_{jck},\Delta^x_{jck})$. We summarize angular and radial structure by
\begin{align}
 m_{jcq}&=\frac{1}{|I_{jc}|}\sum_{k\in I_{jc}}e^{\mathrm{i}q\theta_{jck}},\quad q=1,\ldots,4,\\
 b_{jct}&=\frac{1}{n_{jc}}\sum_k
 \exp\!\left[-\frac{1}{2}\left(\frac{\log(1+r_{jck})-\log(1+t)}{0.30}\right)^2\right],
 \quad t\in\mathcal T,
\end{align}
where $n_{jc}$ is the number of assigned objects, $I_{jc}=\{k:r_{jck}>\epsilon\}$ with $\epsilon=10^{-8}$\,m, and $\mathcal T=\{0,0.5,1,1.5,2,3,4\}$ meters. The category descriptor is
\begin{equation}
\begin{aligned}
h_{jc}=\big[&\mathbf{1}[n_{jc}>0],\log(1+n_{jc}),
\tfrac{1}{n_{jc}}\textstyle\sum_k\mathbf{1}[r_{jck}\leq\epsilon],\\
&\{|m_{jcq}|\}_{q=1}^{4},\{b_{jct}\}_{t\in\mathcal T},
\log(1+\operatorname{median}_k r_{jck})\big].
\end{aligned}
\end{equation}
Empty sets yield zeros. We also encode pairwise angular organization between associated categories $c<c'$ as
\begin{equation}
 h_{jcc'}=\left\{\frac{1}{|I_{jc}|\,|I_{jc'}|}\sum_{k\in I_{jc}}\sum_{l\in I_{jc'}}
 \cos\!\left(q(\theta_{jck}-\theta_{jc'l})\right)\right\}_{q=1}^{4}.
\end{equation}
The room descriptor concatenates anchor presence, the log anchor count, global log counts of associated categories, and the anchor-averaged descriptors:
\begin{equation}
 \phi(x)=\left[\mathbf{1}[N_a>0],\log(1+N_a),
 \{\log(1+N_c)\}_{c\in\mathcal A},
 \frac{1}{\max(N_a,1)}\sum_j\big[\{h_{jc}\}_{c\in\mathcal A},
 \{h_{jcc'}\}_{c<c'}\big]\right].
\end{equation}
The descriptor has four properties. First, every spatial term is measured relative to a functional anchor, so it records how a room organizes itself around the object that gives the room its purpose. Second, only the magnitudes $|m_{jcq}|$ enter, so a room and a rotated copy of it receive identical features. Third, counts enter through $\log(1+n)$ and radii through soft bins, which compresses the range of counts and distances across room sizes. Fourth, an absent category is represented by its presence flag with the remaining entries set to zero, so a missing functional group still contributes to the distance. The descriptor uses object categories and positions only, so it is unaffected by object size and orientation and never sees asset identity, texture, or mesh, which keeps LDD independent of the mesh differences among the asset sources discussed in \cref{app:generation}.

Each reference defines the normalization for every method. Its means and standard deviations, the latter floored at $0.1$, standardize each feature dimension, and the kernel bandwidth is derived from the same reference, so all four methods are scored on one scale and none is measured against statistics taken from its own output.

For generated layouts $G=\{g_i\}_{i=1}^{n}$ and reference layouts $R=\{r_j\}_{j=1}^{m}$, with $x_i=\phi(g_i)$ and $y_j=\phi(r_j)$, LDD is the unbiased maximum mean discrepancy~\citep{gretton2012mmd}:
\begin{align}
\operatorname{LDD}(G,R)
&=\widehat{\operatorname{MMD}}_{k_{\mathrm{RBF}}}^{2}(X,Y),\\
k_{\mathrm{RBF}}(x,y)
&=\exp\!\left(-\frac{\lVert x-y\rVert_2^2}{2\sigma^2}\right),\\
\widehat{\operatorname{MMD}}^{2}(X,Y)
&=\frac{\sum_{i\ne i'}k(x_i,x_{i'})}{n(n-1)}
 +\frac{\sum_{j\ne j'}k(y_j,y_{j'})}{m(m-1)}
 -\frac{2\sum_{i,j}k(x_i,y_j)}{nm}.
\end{align}
The RBF kernel is characteristic, so a zero population MMD implies equality of the represented distributions. LDD is therefore sensitive to differences in typical layouts and to differences in higher-order distribution shape. We set $\sigma$ to the median pairwise distance among standardized reference features and average room-level scores equally across bedrooms, dining rooms, and living rooms. Lower is closer, and the unbiased finite-sample estimate may be slightly negative. Each method contributes ten scenes per room type. 

Kernel Inception Distance (KID) is itself an unbiased MMD estimator with a cubic polynomial kernel applied to Inception image features~\citep{binkowski2018kid}. Our diagnostic Layout-KID keeps the estimator and kernel,
\begin{equation}
 k_{\mathrm{cubic}}(x,y)=\left(1+\frac{x^\top y}{d}\right)^3,
\end{equation}
where $d$ is the feature dimension, but replaces Inception features with the same structured layout representation. It therefore offers a familiar moment-based sensitivity check. LDD stays the primary measure because the RBF kernel tests equality of the represented layout distributions more directly. Multi-scale LDD averages RBF kernels with bandwidths $0.5\sigma$, $\sigma$, and $2\sigma$ to check that conclusions do not depend on one bandwidth.

\subsection{Physical Plausibility}
We follow SceneEval~\citep{sceneeval2026} for method-independent geometric evaluation. Object collision is the percentage of objects participating in at least one mesh intersection, computed per scene and averaged over scenes, and scene collision is the percentage of scenes containing any such object; contact tolerance is zero. OOB is the mean number of objects per scene whose footprint falls outside the room under SceneEval's ray-cast test with a 1\% tolerance.

Dynamic stability follows REST3D~\citep{rest3d2026}. We replay every nonempty exported scene in Isaac Gym~\citep{makoviychuk2021isaacgym} at $1/60$ seconds, using six position iterations, one velocity iteration, and a maximum depenetration velocity of $5$\,m/s. Scenes settle for 120 steps. An object is stable when its final translation is at most $0.1$\,m and its rotation is at most $0.1$\,rad. Stab.\ pools all movable objects; position drift and peak linear and angular velocities provide severity measures. Every method uses the same simulator, timestep, thresholds, coordinate conversion, and evaluation code.

\subsection{Visual Quality}
We use GPT-4o-2024-08-06 to score realism, functionality, layout coherence, and completeness on a 0–10 scale, following the four axes used by SceneWeaver~\citep{sceneweaver2025} and SAGE~\citep{sage2026}. The judge receives a top-down render and a text list of object positions, orientations, and sizes. We use temperature zero and average five judgments per scene, and all four methods are scored in one run. The evaluation interface and score bands are given in Appendix~\ref{app:prompts}.

\section{Additional Scene Generation Results}
\label{app:results}

\subsection{Physical Stability}
\Cref{tab:app-stability} complements the stable-object rate (Stab.) in the main paper with continuous motion severity. Our scenes have the lowest mean position drift and the lowest peak linear and angular velocities under the shared dynamic replay.
\input{tables/app_stability}

\subsection{Real-Layout Distribution Alignment}
\label{app:distribution}\label{app:scan2cad}

\paragraph{References.}
Matterport3D contributes, through its M3DLayout annotations, 236 bedroom, 62 dining-room, and 76
living-room regions. Scan2CAD~\citep{avetisyan2019scan2cad} aligns CAD models to $1{,}506$
ScanNet~\citep{dai2017scannet} scans of 704 distinct spaces. Scan2CAD does not distribute room
types, so we type its rooms with the anchor rules of \cref{app:metrics}, which gives 275
bedrooms, 495 dining rooms, and 287 living rooms from 126, 275, and 152 spaces; applied to
Matterport3D, the same rule leaves the ordering of methods unchanged. Scan2CAD's ShapeNet~\citep{chang2015shapenet}
categories are mapped to our taxonomy, and the mapping is checked against object footprints.

\paragraph{Calibration.}
A real-versus-real control sets the zero point of each reference: ten real rooms from held-out
buildings or spaces, scored against the rest of their reference, give a median LDD of $-0.001$
on Matterport3D and $0.014$ on Scan2CAD over 200 draws, far below every generated LDD.

\paragraph{Results.}
\Cref{tab:app-ldd} adds two kernels to the Matterport3D LDD of \cref{tab:quality}. Under all three, our
layouts are the closest to Matterport3D of all four methods.
\input{tables/app_ldd_references}

\paragraph{Separation from the retrieval corpus.}
Retrieved exemplars do not overlap with the evaluation references, and our layouts remain nearly as far from them as from other exemplars.
None of the thirty scored scenes retrieved a Matterport3D region.
In the feature space of LDD, a scored scene lies on average $6.69$ from the three exemplars it
retrieved and $6.96$ from the other exemplars retrieved for the same room type, only slightly
closer to the layouts it was conditioned on.
Appendix~\ref{app:ablations} examines how retrieval shapes zoning and scene quality.

\subsection{Cost and Efficiency}
Visual proposal generation takes the largest share of our layout time (45.4\%, \cref{tab:app-ours-time}), followed by functional zoning (27.6\%), while lifting with contract-guided compilation (14.5\%) and wall and surface completion (12.4\%) together take under a minute. \Cref{tab:app-resource-time} lists the wall clock behind the generation times of \cref{tab:quality}.

In language-model usage (\cref{tab:app-resource-cost}), our pipeline spends 34,867 tokens and \$0.0057 per scene, 8 times fewer tokens than SAGE and 4 times fewer than SceneWeaver; only Holodeck uses fewer. SAGE's runtime is dominated by its placement loop, as placing objects into rooms takes 54.2 of its 61.3 total generation hours and 407 of its 504 tool calls. Of 392 placement calls with recorded outcomes, 75 place every requested object, 221 partially succeed, and 96 place none. This insertion-heavy trajectory explains its high object density and long runtime.
\input{tables/app_resource_time}
\input{tables/app_ours_time_breakdown}
\input{tables/app_resource_cost}

\section{Ablation Details}
\label{app:ablations}

\subsection{Retrieval}
Retrieval improves how the VLM plans each room and, through it, the quality of the final
scene (\cref{tab:app-retrieval}). We rerun the pipeline from retrieval onward with no
exemplar supplied, which changes only the VLM's input. Without exemplars, the VLM plans
fewer functional zones in four of the five room types, $3.14$ against $3.94$ on average, most
sharply in bathrooms, from $4.40$ to $1.80$, and in dining rooms, from $2.90$ to $1.90$. The
resulting scenes score lower on all four visual-quality axes, most in functionality, $5.25$
against $6.28$, and completeness, $5.87$ against $6.76$, and lie farther from both real
references, especially Scan2CAD, $.2171$ against $.1340$. Retrieval thus informs how the VLM
decomposes each room into functional zones, and the rest of the pipeline builds on that
decomposition.
\input{tables/app_retrieval_metrics}

\subsection{Proposal Source and Contract-Guided Compilation}

\paragraph{Design.}
Both proposal sources receive the same prompt, zoning plan, and asset set, and differ only
in whether a VLM states each object's placement (textual) or the diffusion proposal of our
pipeline supplies it (visual).
For textual proposals, the contract consists of the relations the VLM states along with these placements, drawn from the same relation set and read by the same parser.
Without the contract, the compiler separates objects by geometry alone and skips relational refinement.

\paragraph{Results.}
\Cref{tab:proposal-compiler-ablation} lists the values plotted in
\cref{fig:proposal-compiler-ablation} together with the contract violation rate.
\input{tables/proposal_compiler_ablation}

\section{Object-Goal Navigation Learning}
\label{app:training}

\subsection{Training Sources}
\label{app:nav-sources}
\emph{SceneBloom (ours)} uses the visual-proposal and contract-guided compiler pipeline of
\cref{app:proposal-details} and \cref{app:contract-details}.
\emph{Randomized} starts from those same scenes and keeps every object and mesh, then resamples
each floor object's position uniformly within the room and its yaw uniformly in $[0,2\pi)$,
rejecting samples whose footprint leaves the room or overlaps a floor object already placed.
Surface objects move rigidly with the furniture that supports them and wall objects do not move,
so this source removes the arrangement while holding the object content fixed.
\emph{Agent-planned} replaces the whole placement stage with a single text-only call to
Qwen3-VL-8B-Instruct at temperature zero. The call receives the same room box, the labels and
extents of the same functional zones, and the room's asset library with nominal dimensions, and
returns a textual layout that gives every object's position and yaw and, for wall and surface
objects, the wall or support it belongs to.

\subsection{Navigable Space in the Training Scenes}
\label{app:navmesh}
All pools are measured with one agent geometry (radius $0.17$\,m, height $1.41$\,m, maximum
climb $0.1$\,m, cell height $0.05$\,m). A scene is fragmented when its navigation mesh has more
than one walkable island, and the floor-only rate (floor splits) sets aside raised islands, which
we attribute to the furniture beneath them by sampling each island's height. Walk is the mean over
scenes of the largest island's share of the walkable area. Our layouts are the least fragmented
and have the highest Walk of all six pools (\cref{tab:app-navmesh-walk}).
The same holds for the multi-room houses of \cref{sec:scratch}: across the 80-house training sets, the largest walkable island covers $98.4\%$ of the walkable area in our houses and $86.7\%$ in Randomized houses.

Fragmentation carries into training because episode construction admits a fragmented scene: it
requires a finite geodesic between start and goal, so each episode stays on one island. For the
randomized control, $41.1\%$ of training episodes come from fragmented scenes, against $2.0\%$ for
ours, and these episodes are shorter, with a median geodesic length of $1.94$\,m against $2.27$\,m
in its connected scenes.

\input{tables/app_navmesh_walk}

\subsection{Training from Scratch}
\label{app:scratch}

\paragraph{Houses and episodes.}
Randomized houses resample the floor objects within each room of our houses as in \cref{app:nav-sources}, Agent-planned houses are assembled in the same way from rooms laid out by the text agent, and real scans are HM3D training scenes.
All sources draw episodes with the same start-distance sampler.
The STOP rates of \cref{sec:scratch} are measured with 80 houses and average three training seeds.
Grouped by geodesic distance from the start to the nearest goal viewpoint, the rate in our houses is 94.7\% below 2\,m, 94.6\% for 2--5\,m, and 88.7\% from 5\,m on, against 85.4\%, 82.1\%, and 65.9\% in real scans.

\paragraph{Training and evaluation.}
Each episode ends at STOP or after 500 steps. Success follows SemExp~\citep{chaplot2020semantic} and the Habitat ObjectNav challenge it adopts, which require calling STOP within 1\,m of a goal object at a location from which the object is visible.
During training, the goal policy reads semantic labels from the simulator.
Evaluation uses the 1,000 validation episodes of the HM3D ObjectNav v2 dataset.
The untrained policy is the same goal policy at its random initialization.
\cref{tab:app-scratch} lists SR and SPL at each house count of \cref{fig:scratch}.

\input{tables/app_semexp_scratch}

\subsection{Fine-Tuning a Pretrained Policy}
\label{app:finetune}

\paragraph{Training recipe.}
Every fine-tuning run, including real-scan continuation, scales the learning rate of the state
encoder by 0.1.

\paragraph{Training episodes.}
For each scene, we request up to four episodes per available goal category. Habitat samples a navigable
start, which is retained only when the nearest reachable visible goal viewpoint is 1.5--12\,m away
geodesically.

\paragraph{Perception conditions.}
Fine-tuning on every source, including the real scans, uses ground-truth target masks, and every evaluation of a fine-tuned policy uses
masks predicted by YOLOE~\citep{wang2025yoloe}, so the deployed perception interface is the predicted mask.

\paragraph{Evaluation episodes.}
We select 24 episodes in each of the 36 scenes of HM3D-OVON Val Unseen, sampling in round-robin
order across the goal categories available in each scene.

\subsection{Further Evidence on Fine-Tuning}
\label{app:finetune-evidence}

\paragraph{Cleaned real episodes.}
On real episodes cleaned of annotation errors, the gains from our scenes grow, and SceneBloom with occluders achieves the highest SR and SPL (\cref{tab:app-verified}).
Some of the 864 real evaluation episodes have goal annotations that do not match the object, and 86 of them have no correctly annotated goal, so a policy cannot succeed on them by finding the requested object.
We inspect every annotated goal instance in renderings from its goal viewpoints and remove each episode whose goals include an instance that depicts another category (e.g., a mirror labeled as a picture), an object outside the meaning of its label (e.g., a wine cooler labeled as a refrigerator), text-only wall decoration labeled as a picture, or no identifiable object.
This filter is fixed before any policy is scored and keeps 631 episodes covering all 36 scenes, on which we re-score the evaluation runs of \cref{tab:learning,tab:occlusion}.
On these episodes, the SR gain of SceneBloom over the pretrained policy grows from 0.31 to 0.52 points, and SceneBloom also leads in SPL.
The occluder zone adds more, as SceneBloom with occluders exceeds SceneBloom by 1.48 points and the pretrained policy by 2.00 points in SR, against 0.93 and 1.24 points on all 864 episodes.
Removing annotation errors therefore makes the gains from our scenes clearer.

\paragraph{Search difficulty.}
The occluder zone brings the search difficulty of our training episodes closer to that of the real evaluation episodes (\cref{tab:app-difficulty}).
We measure this difficulty without any policy, as the share of episodes whose goal stays hidden after a full turn in place at the start position and the share whose goal is hidden at the start heading.
Adding occluders raises these shares from 0.8\% to 16.4\% and from 50.1\% to 68.2\%, against 58.9\% and 87.7\% in the real episodes.
Randomized with occluders is even harder to search (18.4\% and 81.5\%), yet it transfers worse than SceneBloom with occluders on both the full and the cleaned real episodes (\cref{tab:occlusion,tab:app-verified}).
Difficulty therefore improves transfer when it is embedded in a coherent layout.

\input{tables/app_finetune_evidence}

\section{Qualitative Results}
\begin{figure}[h]
    \centering
    \includegraphics[width=\linewidth]{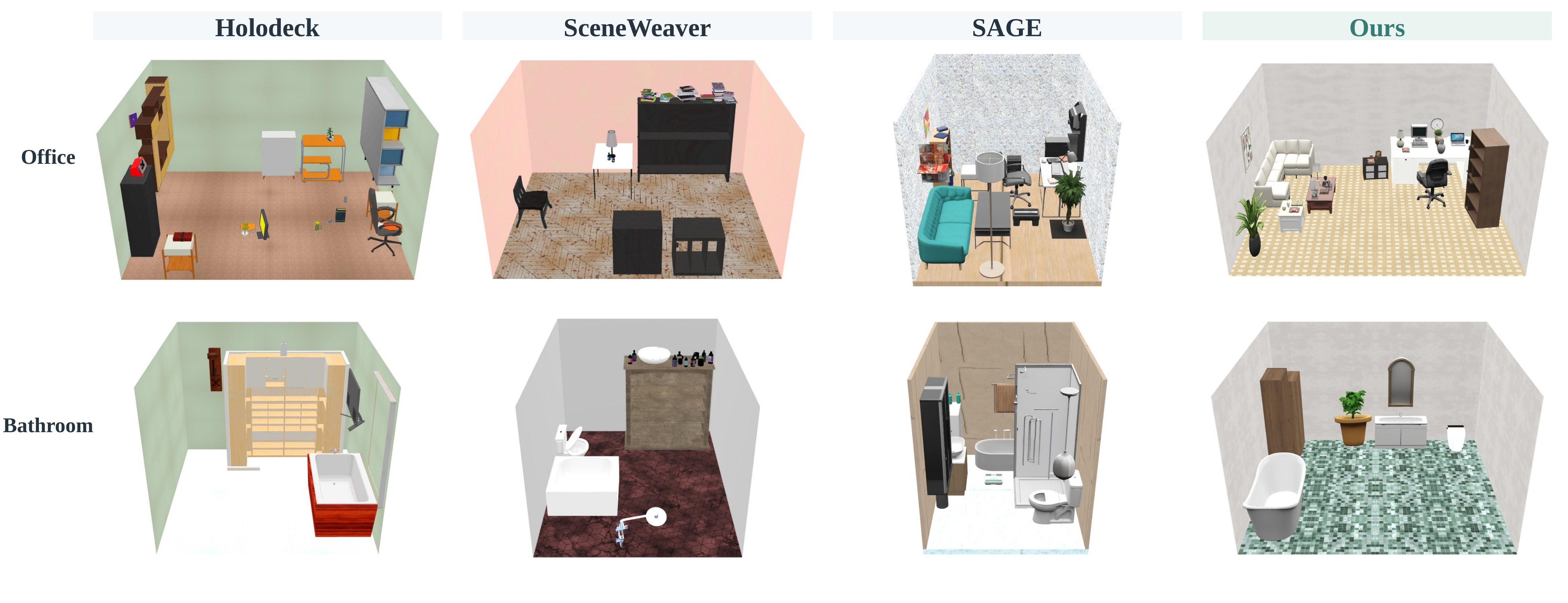}
    \caption{
        \textbf{Additional qualitative comparisons on offices and bathrooms.}
    }
    \label{fig:qualitative-appendix}
\end{figure}

\Cref{fig:qualitative-appendix} extends the main comparison to offices and bathrooms.

\input{sections/A_prompts}

%% file: tables/app_contract_relations.tex
\begin{table}[h]
\centering\small
\caption{Relations a zone contract can hold. The VLM selects relations; the compiler checks them
geometrically and performs all pose changes.}
\label{tab:app-contract-relations}
\begin{tabular}{lll}
\toprule
Relation & Meaning & Permitted refinement \\
\midrule
\texttt{face} & Subject faces the target. & Yaw, or moving in front of it. \\
\texttt{align\_with} & Subject is axis-aligned with the target. & Yaw. \\
\texttt{beside} & Subject stands at the target's side. & Yaw, or moving to its side. \\
\texttt{move\_near} & Subject stays within reach of the target. & Moving and turning toward it. \\
\texttt{back\_to} & Subject's back is against its zone's wall. & Yaw, or moving to the wall. \\
\bottomrule
\end{tabular}
\end{table}

%% file: tables/app_stability.tex
\begin{table}[h]
  \centering
  \caption{\textbf{Physical stability under the REST3D protocol~\citep{rest3d2026}.}
  Stab.\ is the percentage of movable objects within the translation and rotation thresholds,
  and the last three columns measure motion severity.
  Bold/underline mark best/second-best.}
  \label{tab:app-stability}
  \small
  \begin{tabular}{lcccc}
    \toprule
    Method & Stab.\ $\uparrow$ & Pos. Drift (m) $\downarrow$ & Lin. Vel. (m/s) $\downarrow$ & Ang. Vel. (rad/s) $\downarrow$ \\
    \midrule
    Holodeck & 35.1 & 0.7923 & 2.573 & 16.727 \\
    SceneWeaver & \underline{45.3} & \underline{0.3698} & \underline{1.974} & 14.219 \\
    SAGE & 45.1 & 0.9757 & 2.470 & \underline{11.015} \\
    \midrule
    \rowcolor{oursbg}
    SceneBloom (ours) & \textbf{75.3} & \textbf{0.0996} & \textbf{0.811} & \textbf{4.724} \\
    \bottomrule
  \end{tabular}
\end{table}

%% file: tables/app_ldd_references.tex
\begin{table}[h]
\centering\small
\caption{\textbf{Layout alignment with Matterport3D under three kernels.} LDD is the metric of
\cref{tab:quality}; multi-scale LDD and Layout-KID replace its kernel as defined in
\cref{app:layout-metrics}. Bold/underline mark best/second-best.}
\label{tab:app-ldd}
\begin{tabular}{lccc}
\toprule
Method & LDD $\downarrow$ & Multi-scale LDD $\downarrow$ & Layout-KID $\downarrow$ \\
\midrule
Holodeck & .2509 & .2082 & 5.0997 \\
SceneWeaver & \underline{.2009} & \underline{.1784} & \underline{4.8749} \\
SAGE & .2191 & .1948 & 5.3881 \\
\midrule
\rowcolor{oursbg}
SceneBloom (ours) & \textbf{.1759} & \textbf{.1617} & \textbf{2.0391} \\
\bottomrule
\end{tabular}
\end{table}

%% file: tables/app_resource_time.tex
\begin{table}[h]
\centering
\caption{\textbf{Generation wall clock per scene.} Measured from each run's own artifacts, one scene
at a time, over the baseline pools of \cref{tab:quality} and a 50-scene run of our pipeline. The span is layout generation for every system:
rendering and physical settling are excluded. Each job generates one scene on one H100-class GPU,
except SceneWeaver, which runs on 8 CPU cores. Bold/underline mark best/second-best.}
\label{tab:app-resource-time}
\small
\begin{tabular}{lrr}
\toprule
Method & Mean (min) $\downarrow$ & Median (min) $\downarrow$ \\
\midrule
Holodeck \citep{holodeck2024}       & \phantom{00}\textbf{1.6} & \phantom{00}\textbf{1.5} \\
SceneWeaver \citep{sceneweaver2025} & 313.0           & 298.8 \\
SAGE \citep{sage2026}               & \phantom{0}73.6 & \phantom{0}77.9 \\
\midrule
\rowcolor{oursbg}
SceneBloom (ours)                   & \phantom{00}\underline{2.7} & \phantom{00}\underline{2.6} \\
\bottomrule
\end{tabular}
\end{table}

%% file: tables/app_ours_time_breakdown.tex
\begin{table}[h]
\centering
\caption{\textbf{Our layout-generation time breakdown.} Mean wall clock over a 50-scene production run with the same scene requests as \cref{tab:quality}. Components differ from the total by 0.2 seconds due to uninstrumented orchestration and rounding.}
\label{tab:app-ours-time}
\small
\begin{tabular}{lrr}
\toprule
Stage & Mean (s) & Share \\
\midrule
Functional zoning & 45.3 & 27.6\% \\
Visual proposal generation & 74.6 & 45.4\% \\
Lifting and contract-guided compilation & 23.8 & 14.5\% \\
Wall and surface completion & 20.3 & 12.4\% \\
\midrule
Layout generation & 164.2 & 100.0\% \\
\bottomrule
\end{tabular}
\end{table}

%% file: tables/app_resource_cost.tex
\begin{table}[h]
\centering
\caption{\textbf{Language-model usage and cost per scene.} Bold/underline mark best/second-best.}
\label{tab:app-resource-cost}
\small
\setlength{\tabcolsep}{4pt}
\begin{tabular}{llrr}
\toprule
Method & LLM backbone & Tokens/scene $\downarrow$ & USD/scene $\downarrow$ \\
\midrule
Holodeck        & Qwen3-VL-8B-Instruct & \phantom{0,00}\textbf{6,986} & \textbf{\$0.0016} \\
SceneWeaver & Qwen3-VL-8B-Instruct & \phantom{0,}154,649 & \$0.0228 \\
SAGE                & GPT-OSS-120B $+$ Qwen3-VL-30B-A3B-Instruct & \phantom{0,}293,499 & \$0.0151 \\
\midrule
\rowcolor{oursbg}
SceneBloom (ours)                   & Qwen3-VL-8B-Instruct & \phantom{0,0}\underline{34,867} & \underline{\$0.0057} \\
\bottomrule
\end{tabular}
\end{table}

%% file: tables/app_retrieval_metrics.tex
\begin{table}[h]
\centering
\caption{\textbf{Retrieval ablation.} Results on 50 scenes, ten per room type; LDD uses the 30 bedroom, dining-room, and living-room scenes. Zones is the mean
number of functional zones per scene, and the visual quality of both rows is judged in a single
run, so the with-retrieval row rescores the scenes of \cref{tab:quality} and differs from it by at most 0.02. Bold marks the better value.}
\label{tab:app-retrieval}
\small
\begin{tabular}{c c cccc cc}
\toprule
& & \multicolumn{4}{c}{Visual Quality $\uparrow$} & \multicolumn{2}{c}{LDD $\downarrow$} \\
\cmidrule(lr){3-6}\cmidrule(lr){7-8}
Retrieval & Zones & Real. & Func. & Coh. & Compl. & MP3D & Scan2CAD \\
\midrule
\xmark & 3.14 & 5.89 & 5.25 & 4.63 & 5.87 & .1857 & .2171 \\
\rowcolor{oursbg}
\cmark & 3.94 & \textbf{6.19} & \textbf{6.28} & \textbf{4.75} & \textbf{6.76} & \textbf{.1759} & \textbf{.1340} \\
\bottomrule
\end{tabular}
\end{table}

%% file: tables/proposal_compiler_ablation.tex
\begin{table}[t]
\caption{\textbf{Ablation of proposal source and contract guidance.} Values plotted in \cref{fig:proposal-compiler-ablation}. Viol.\ is the percentage of relations in the extracted contract that the compiled layout violates. Bold/underline mark best/second-best.}
\label{tab:proposal-compiler-ablation}
\centering
\small
\setlength{\tabcolsep}{4pt}
\begin{tabular}{lc ccc cccc c}
\toprule
& & \multicolumn{3}{c}{Physical Plausibility} & \multicolumn{4}{c}{Visual Quality} & Relations \\
\cmidrule(lr){3-5}\cmidrule(lr){6-9}\cmidrule(lr){10-10}
Proposal & Contract & Coll.\ obj $\downarrow$ & Coll.\ scene $\downarrow$ & Stab.\ $\uparrow$ & Real.\ $\uparrow$ & Func.\ $\uparrow$ & Coh.\ $\uparrow$ & Compl.\ $\uparrow$ & Viol.\ $\downarrow$ \\
\midrule
Textual & \xmark & 0.67 & 4.00 & 98.44 & 5.44 & 5.47 & 4.29 & 3.80 & 39.07 \\
Textual & \cmark & 2.00 & 8.00 & 94.27 & 5.50 & 5.59 & \underline{4.37} & 3.93 & \underline{10.75} \\
\midrule
Visual & \xmark & \textbf{0.00} & \textbf{0.00} & \textbf{100.00} & \underline{5.66} & \underline{5.65} & 4.31 & \underline{4.05} & 37.45 \\
\rowcolor{oursbg}
Visual & \cmark & \textbf{0.00} & \textbf{0.00} & \underline{99.74} & \textbf{5.83} & \textbf{5.85} & \textbf{4.65} & \textbf{4.07} & \textbf{4.25} \\
\bottomrule
\end{tabular}
\end{table}

%% file: tables/app_navmesh_walk.tex
\begin{table}[t]
\centering\small
\caption{\textbf{Navigation-mesh connectivity.} Our
layouts and the two controls are measured on their fine-tuning pools of \cref{sec:finetune}, the baselines on their pools
of \cref{tab:quality}. Bold/underline mark best/second-best.}
\label{tab:app-navmesh-walk}
\begin{tabular}{lrrr}
\toprule
Scene pool & Fragmented (\%) $\downarrow$ & Floor splits (\%) $\downarrow$ & Walk $\uparrow$ \\
\midrule
Holodeck & 16.0 & \underline{2.0} & \underline{0.9840} \\
SAGE & 20.0 & 11.1 & 0.9467 \\
SceneWeaver & 20.0 & 12.0 & 0.9506 \\
\midrule
Randomized & 37.9 & 36.8 & 0.8823 \\
Agent-planned & \underline{14.2} & 14.2 & 0.9663 \\
\midrule
\rowcolor{oursbg}
SceneBloom (ours) & \textbf{2.6} & \textbf{0.9} & \textbf{0.9968} \\
\bottomrule
\end{tabular}
\end{table}

%% file: tables/app_semexp_scratch.tex
\begin{table}[h]
  \centering
  \caption{\textbf{Training from scratch.} SR and SPL (\%) on the 36 real HM3D validation homes for 40, 80,
  and 120 training houses. Trained rows average three training seeds. Bold/underline mark best/second-best
  at each house count.}
  \label{tab:app-scratch}
  \small
  \begin{tabularx}{\linewidth}{l*{6}{Y}}
  \toprule
  & \multicolumn{3}{c}{SR} & \multicolumn{3}{c}{SPL} \\
  \cmidrule(lr){2-4}\cmidrule(l){5-7}
  Training source & 40 & 80 & 120 & 40 & 80 & 120 \\
  \midrule
  Untrained & \multicolumn{3}{c}{29.70} & \multicolumn{3}{c}{12.16} \\
  \midrule
  Real scans & 33.70 & 32.53 & 32.87 & 15.77 & 15.42 & 15.43 \\
  Randomized & 32.53 & 32.67 & 33.50 & 15.24 & 15.66 & \underline{16.04} \\
  Agent-planned & \underline{34.00} & \underline{34.07} & \underline{34.40} & \textbf{16.14} & \underline{15.73} & 15.85 \\
  \midrule
  \rowcolor{oursbg}
  SceneBloom (ours) & \textbf{34.70} & \textbf{35.57} & \textbf{35.63} & \underline{15.83} & \textbf{16.25} & \textbf{16.20} \\
  \bottomrule
  \end{tabularx}
\end{table}

%% file: tables/app_finetune_evidence.tex
\begin{table}[h]
  \centering
  \begin{minipage}[b]{0.45\linewidth}
  \caption{\textbf{Real-environment transfer on cleaned episodes.} SR and SPL (\%) on the 631
  HM3D-OVON episodes without annotation errors.}
  \label{tab:app-verified}
  \end{minipage}\hfill
  \begin{minipage}[b]{0.53\linewidth}
  \caption{\textbf{Search difficulty of navigation episodes.} Share of episodes (\%) whose goal stays
  hidden after a full turn at the start position or is hidden at the start heading, measured
  with one camera for every set.}
  \label{tab:app-difficulty}
  \end{minipage}
  \par\vspace{-11pt}
  \begin{minipage}[t]{0.45\linewidth}
  \vspace{0pt}
  \centering
  \small
  \setlength{\tabcolsep}{4pt}
  \begin{tabular}{lcc}
  \toprule
  Policy & SR & SPL \\
  \midrule
  Pretrained & 49.45 & 23.09 \\
  Real-scan continuation & 49.87 & 22.78 \\
  \midrule
  \multicolumn{3}{l}{\textit{Without occluders}} \\
  Randomized & 49.71 & 22.74 \\
  Agent-planned & 49.45 & 22.65 \\
  SceneBloom & 49.97 & \underline{23.22} \\
  \cmidrule(lr){1-3}
  \multicolumn{3}{l}{\textit{With occluders}} \\
  Randomized & \underline{50.45} & 22.85 \\
  \rowcolor{oursbg}
  SceneBloom & \textbf{51.45} & \textbf{23.63} \\
  \bottomrule
  \end{tabular}
  \end{minipage}\hfill
  \begin{minipage}[t]{0.53\linewidth}
  \vspace{0pt}
  \centering
  \small
  \setlength{\tabcolsep}{4pt}
  \begin{tabular}{lcc}
  \toprule
  Episode set & \shortstack{Hidden after\\a full turn} & \shortstack{Hidden at\\start heading} \\
  \midrule
  \multicolumn{3}{l}{\textit{Without occluders}} \\
  SceneBloom & 0.8 & 50.1 \\
  \cmidrule(lr){1-3}
  \multicolumn{3}{l}{\textit{With occluders}} \\
  Randomized & 18.4 & 81.5 \\
  \rowcolor{oursbg}
  SceneBloom & 16.4 & 68.2 \\
  \midrule
  Real evaluation episodes & 58.9 & 87.7 \\
  \bottomrule
  \end{tabular}
  \end{minipage}
\end{table}

%% file: sections/A_prompts.tex
\section{Prompt Templates}
\label{app:prompts}
This appendix gives, in condensed form, the prompts of three model calls: functional zoning
(\cref{app:prompt-zoner}) and relation-contract extraction (\cref{app:prompt-contract}) in the
generation pipeline, and the visual-quality judge used only for evaluation
(\cref{app:prompt-vlm}). The pipeline also calls the VLM to choose wall and surface objects and
propose their positions (\cref{app:completion}). Bracketed fields are filled for each
call with the inputs named in the box.

\subsection{Functional Zoning}
\label{app:prompt-zoner}
\leavevmode\vspace*{-\dimexpr\baselineskip+\parskip\relax}
\begin{promptbox}{Functional zoning: system message}
You are a furniture-layout expert specializing in indoor scene generation. You receive up to three retrieved indoor floor plans as spatial-layout references. Each reference provides a top-down render when available and a record of its room type, object count, and, for every object, category, ground-plane center and size in meters, height, and yaw. Use the references only as spatial-layout priors and produce a JSON zone plan for the room requested by the user. The zones must be non-overlapping axis-aligned rectangles that tile the room floor.

\medskip
Return only valid JSON with this structure:
\begin{promptjson}
\footnotesize\ttfamily\raggedright
\{"room\_type": STRING, "room\_width\_m": FLOAT, "room\_depth\_m": FLOAT, "zones": [\{"label": STRING, "polygon\_2d": [[x1,z1],\ldots,[x4,z4]], "allowed\_categories": [STRING], "category\_counts": \{STRING: INT\}, "anchor\_category": STRING or null, "max\_objects": INT, "priority": INT, "diffusion\_prompt": STRING, "wall\_orientation": north, south, east, west, or null\}]\}
\end{promptjson}

Apply the following rules:
\begin{enumerate}[leftmargin=1.5em,itemsep=2pt,topsep=3pt]
\item Express all polygons in the local XZ frame, inside $[0,W]\times[0,D]$, where $W$ and $D$ are the room width and depth; zones must cover the floor without overlap or gaps.
\item Infer realistic room dimensions from the references and allocate enough area for the furniture assigned to each zone.
\item Use only the floor-furniture categories listed in the prompt, such as bed, sofa, desk, and dining\_table. Wall-mounted, hanging, built-in, and surface objects are added by later stages.
\item Give each allowed category an integer count, by default 1, and use more only when the room type naturally has multiples. A circulation zone that absorbs leftover floor has no categories, no anchor, and \texttt{max\_objects} of 0.
\item If present, \texttt{anchor\_category} must be one of the zone's allowed categories.
\item Write the zone prompt (\texttt{diffusion\_prompt}) in at most 25 words. Refer to each planned item by its exact category name with no adjective before it, state relative relations between items, and mention no unplanned object or absolute room position.
\item Set wall orientation from the zone boundary: north at $z=D$, south at $z=0$, east at $x=W$, and west at $x=0$. For a corner zone, use the wall the anchor backs onto; use null for an interior or empty zone.
\end{enumerate}
\end{promptbox}

\begin{promptbox}{Functional zoning: user message}
The VLM receives, for each retrieved reference, its top-down render when available, its similarity score, and \texttt{[REFERENCE\_JSON]}; then the available asset catalog, the original room instruction, the room type, and the required categories, followed by: ``Produce the JSON zone plan for this room.'' When a zone's prompt and category counts disagree, a note listing each mismatch is appended and the VLM returns the complete plan again, at most twice.
\end{promptbox}

\subsection{Relation-Contract Extraction}
\label{app:prompt-contract}
\leavevmode\vspace*{-\dimexpr\baselineskip+\parskip\relax}
\begin{promptbox}{Relation-contract extraction: system message}
You are reading the original diffusion-generated image of one room zone. The accompanying JSON lists the reconstructed objects and the identifiers that must be used. List the qualitative spatial relations shown in the image, whether or not the current 3D reconstruction satisfies them, and report only relations that are clearly visible. Do not output coordinates or numeric yaw, and do not invent identifiers. Use only the five relations of \cref{tab:app-contract-relations}; \texttt{face} may be used only for subjects whose front is unambiguous, and \texttt{back\_to} takes only the wall behind the zone as its target. Confidence expresses how clearly the image supports the relation: 0.9--1.0 is unmistakable, 0.7--0.85 is clear, and 0.4--0.6 is plausible but ambiguous.

\medskip
Return at most 10 relations as JSON:
\begin{promptjson}
\footnotesize\ttfamily\raggedright
\{"zone\_assessment": \{"function": STRING\}, "constraints": [\{"subject\_id": ID, "operator": RELATION, "target": \{"type": "object", "id": ID\} or \{"type": "wall"\}, "confidence": FLOAT, "reason": STRING\}]\}
\end{promptjson}
\end{promptbox}

\begin{promptbox}{Relation-contract extraction: user message}
The VLM receives the original zone image, followed by the list of objects in the zone, \texttt{[RECONSTRUCTED\_OBJECT\_JSON]}.
\end{promptbox}

\subsection{Visual-Quality Evaluation}
\label{app:prompt-vlm}
\begin{promptbox}{Visual-quality evaluation: user message}
You are an expert interior-design reviewer scoring a computer-generated 3D \texttt{[ROOM\_TYPE]} scene produced for \texttt{[ROOM\_INSTRUCTION]}. You are shown a top-down render and a text list of every object's position, facing, and size, \texttt{[OBJECT\_LIST]}. Score strictly: if a critical issue exists, including a missing key object, an obvious layout error, an unrealistic element, or a clearly wrong size, lower the score substantially even when other aspects are fine.

\medskip
Score each criterion independently on an integer 0--10 scale, where 10 fully meets or exceeds a professional designer's expectation and needs no major fix, 5 partially meets it with obvious flaws, and 0 completely fails. Use the full range rather than defaulting to its midpoint. For every criterion use the same calibration bands:
\begin{itemize}[leftmargin=1.5em,itemsep=2pt,topsep=3pt]
\item 8--10: the criterion is fully or substantially met; any deviations are minor;
\item 4--7: the criterion is partially met; notable deficiencies are present;
\item 0--3: the criterion is not met; deficiencies are severe.
\end{itemize}
The four criteria are:
\begin{enumerate}[leftmargin=1.5em,itemsep=2pt,topsep=3pt]
\item \textbf{Realism}: how realistic and lived-in the room looks, ignoring texture, lighting, and doors.
\item \textbf{Functionality}: how well the room supports its intended activities, such as sleeping or working.
\item \textbf{Layout coherence}: whether furniture is arranged logically, in a good pose, and well sized.
\item \textbf{Completeness}: how complete and finished the room feels.
\end{enumerate}
Reason briefly for each criterion, then output only:
\begin{promptjson}
\footnotesize\ttfamily\raggedright
\{"realism": \{"score": INT, "justification": STRING\}, "functionality": \{\ldots\}, "layout\_coherence": \{\ldots\}, "completeness": \{\ldots\}\}
\end{promptjson}
\end{promptbox}